\documentclass[10pt,twocolumn,letterpaper]{article}

\usepackage{hyperref}
\usepackage{cleveref}
\usepackage{iccv}

\usepackage{amssymb}
\usepackage{latexsym}
\usepackage{times,multirow}
\usepackage{epsfig}
\usepackage{epstopdf}
\usepackage{amsmath}
\usepackage{amssymb}
\usepackage{scalefnt}
\usepackage{graphicx}
\usepackage[font=small]{caption}
\usepackage[font=footnotesize,caption=false]{subfig}
\usepackage{multirow}
\usepackage{enumitem}
\usepackage{threeparttable}
\usepackage{gensymb}
\usepackage{capt-of}
\usepackage{relsize}
\usepackage{color}
\usepackage{mathpazo}
\usepackage{url}
\usepackage{xcolor}
\definecolor{newcolor}{rgb}{.8,.349,.1}

\iccvfinalcopy 

\makeatletter
\newcommand{\clearsubcaptcounter}{\setcounter{sub\@captype}{0}}
\makeatother

\begin{document}

\title{Spatio-temporal interaction model for crowd video analysis}

\author{Neha Bhargava\\
Indian Institute of Technology Bombay\\
India\\
{\tt \small neha@ee.iitb.ac.in}
\and
Subhasis Chaudhuri\\
Indian Institute of Technology Bombay\\
India\\
{\tt\small sc@ee.iitb.ac.in}
}

\maketitle

\begin{abstract}
We present an unsupervised approach to analyze crowd at various levels of granularity $-$ individual, group and collective. We also propose a motion model to represent the collective motion of the crowd. The model captures the spatio-temporal interaction pattern of the crowd from the trajectory data captured over a time period. Furthermore, we also propose an effective group detection algorithm that utilizes the eigenvectors of the interaction matrix of the model. We also show that the eigenvalues of the interaction matrix characterize various group activities such as being stationary, walking, splitting and approaching. The algorithm is also extended trivially to recognize individual activity. Finally, we discover the overall crowd behavior by classifying a crowd video in one of the eight categories. Since the crowd behavior is determined by its constituent groups, we demonstrate the usefulness of group level features during classification. Extensive experimentation on various datasets demonstrates a superlative performance of our algorithms over the state-of-the-art methods.
\end{abstract}

\maketitle


Understanding human behavior at an individual level, at a group level and at a crowd level in different scenarios has always attracted the researchers. The variability and complexity in the behavior make it a highly challenging task. However, this decade is witnessing a huge interest of researchers in the area of crowd motion analysis due to its various applications in surveillance, safety, public place management, hazard prevention, and virtual environments. This interest has resulted in many interesting papers in the area. We are aware of at least four survey papers on the subject of crowd analysis  that indicate the amount of attention, it has drawn in this and the previous decade \cite{survey1},\cite{survey2},\cite{survey3},\cite{survey4}. The latest survey paper \cite{survey1} by Chang \etal encapsulates the recent works published after 2009, covering topics of motion pattern segmentation, crowd behavior and anomaly detection. Thida \etal \cite{survey2} provide a review on macroscopic and microscopic modeling methods. They also present a critical survey on crowd event detection. Julio \etal cover various vision techniques applicable to crowd analysis such as tracking, density estimation, and computer simulation \cite{survey3}. Zhan \textit{et al.} discuss various vision based techniques used in crowd analysis. They also discuss crowd analysis from the perspective of different disciplines $-$ psychology, sociology and computer graphics \cite{survey4}. At the top level, the techniques used in crowd motion analysis can be divided into two major classes $-$ holistic and particle based. The holistic methods consider crowd as a single entity and analyze the overall behavior. These methods fail to provide much insight at an individual or intermediate level. On the other hand, particle based methods consider crowd as a collection of individuals. But their performance degrades with the increase in crowd density due to occlusion and tracking problems. The analysis at intermediate level \ie~at group level might provide more insights at individual and overall levels.

\begin{figure*}
\centering
	\subfloat[\label{fig:crowd2}Uniform crowd]{\includegraphics[trim={0.5cm 0 0 0},clip, scale=0.28]{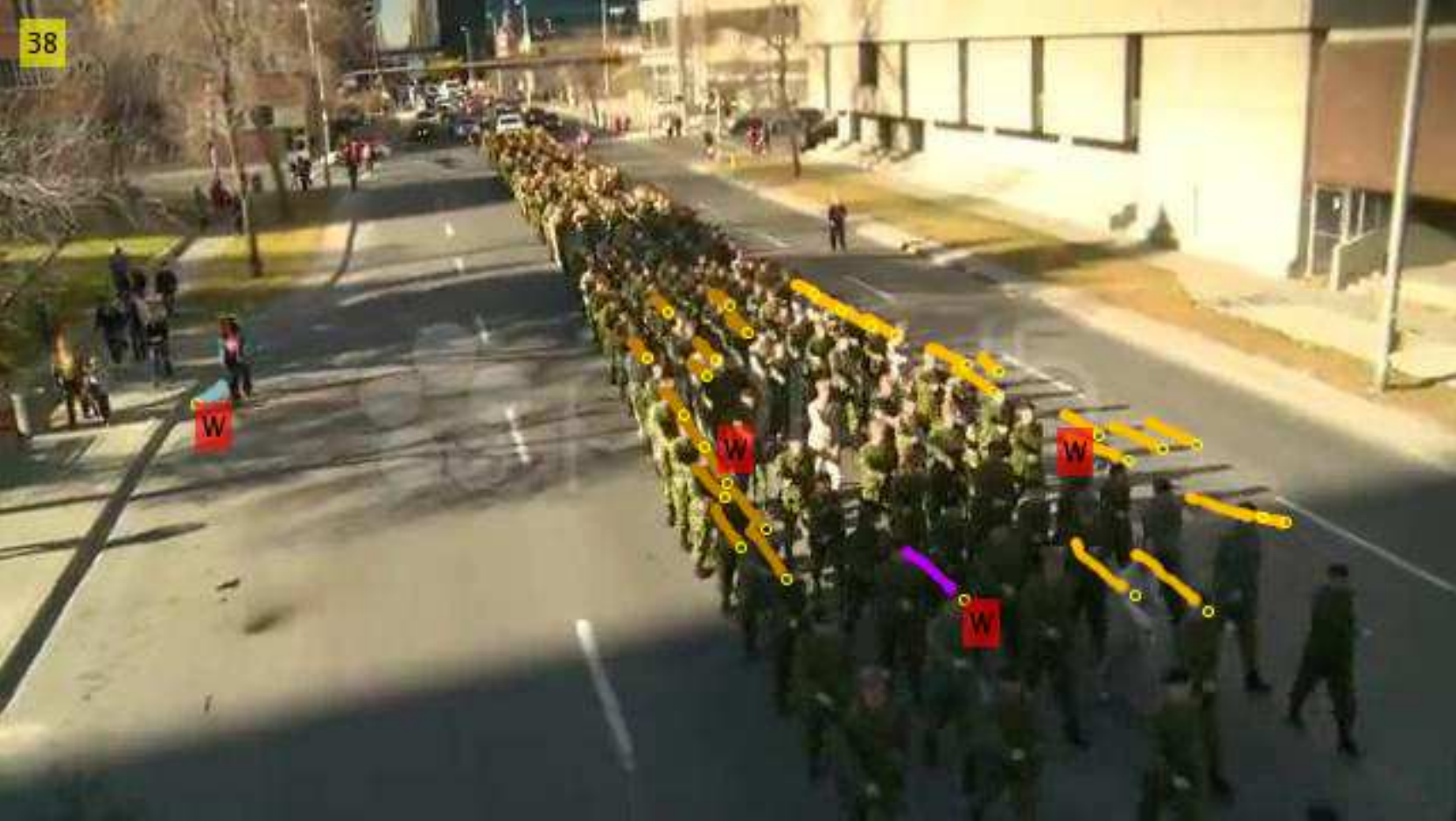}}~
\subfloat[\label{fig:crowd1}Mixed crowd]{\includegraphics[trim={0 1.25cm 0 0cm},clip, scale=0.195]{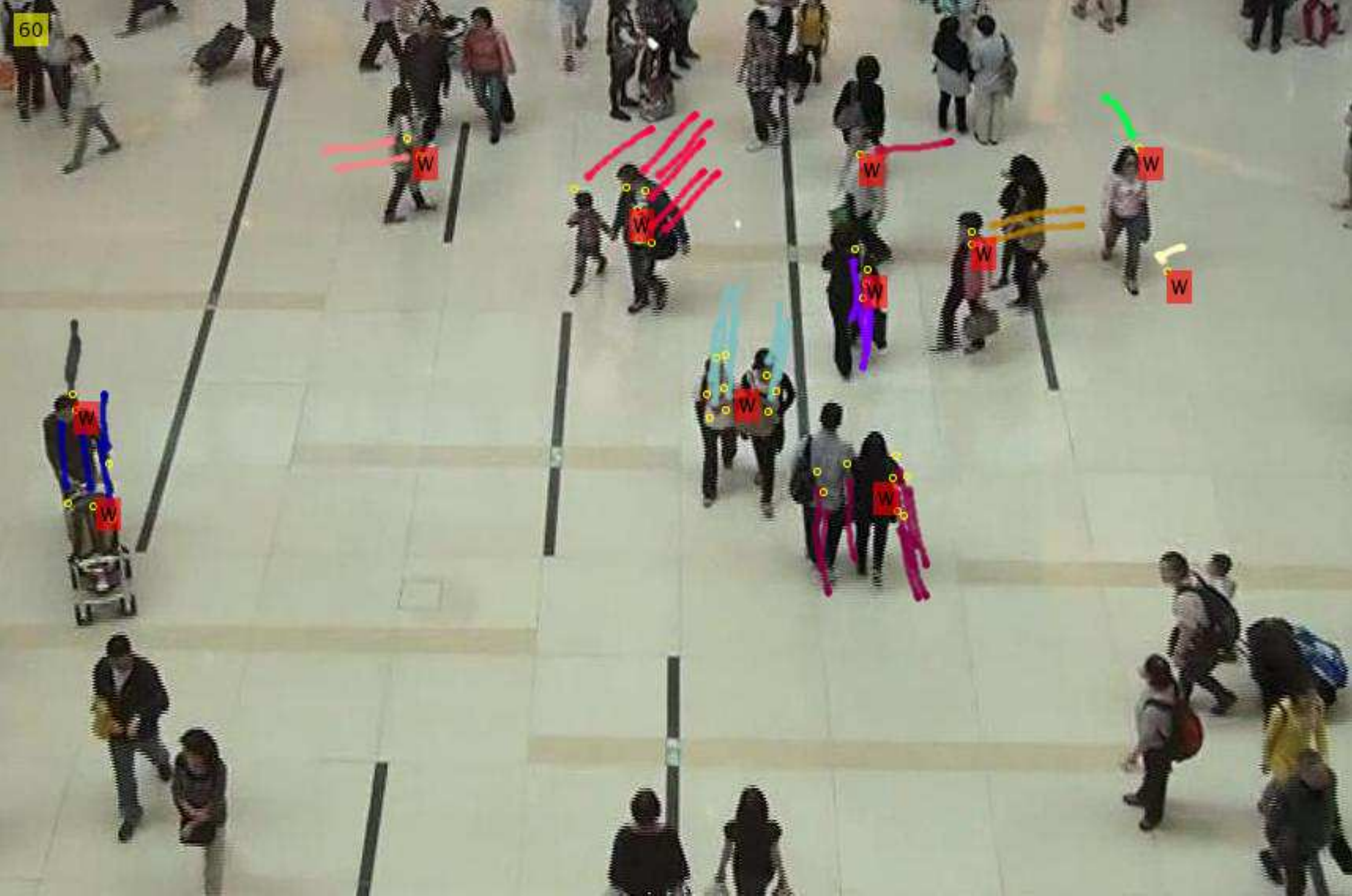}}~
	\subfloat[\label{fig:stand} Stationary group]{\includegraphics[trim={3.5cm 5.25cm 1.8cm 0.5cm},clip, scale=0.4]{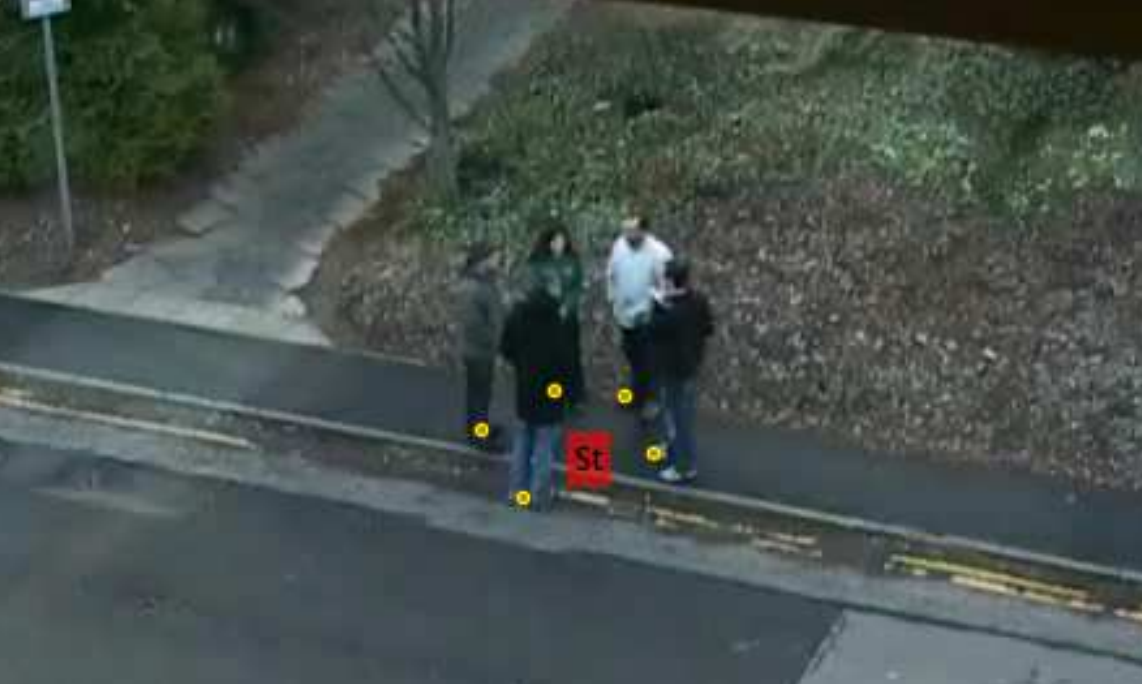}}\\ 
	\subfloat[\label{fig:walk}Walking]{\includegraphics[trim={18cm 5.75cm 10cm 5.7cm},clip, scale=0.2]{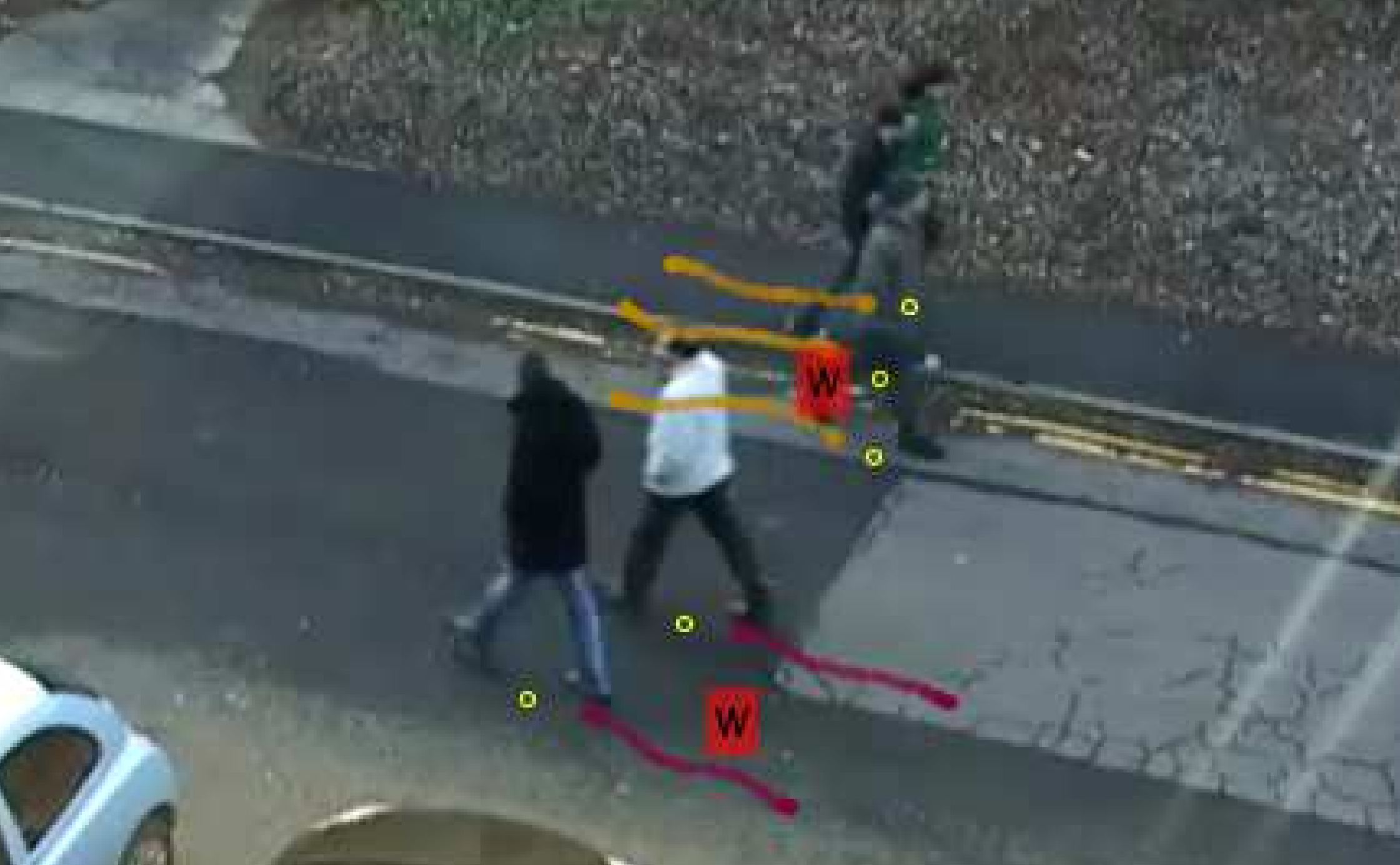}}~
	\subfloat[\label{fig:merge}Approaching]{\includegraphics[trim={3.5cm 5.15cm 1.8cm 0.5cm},clip, scale=0.4]{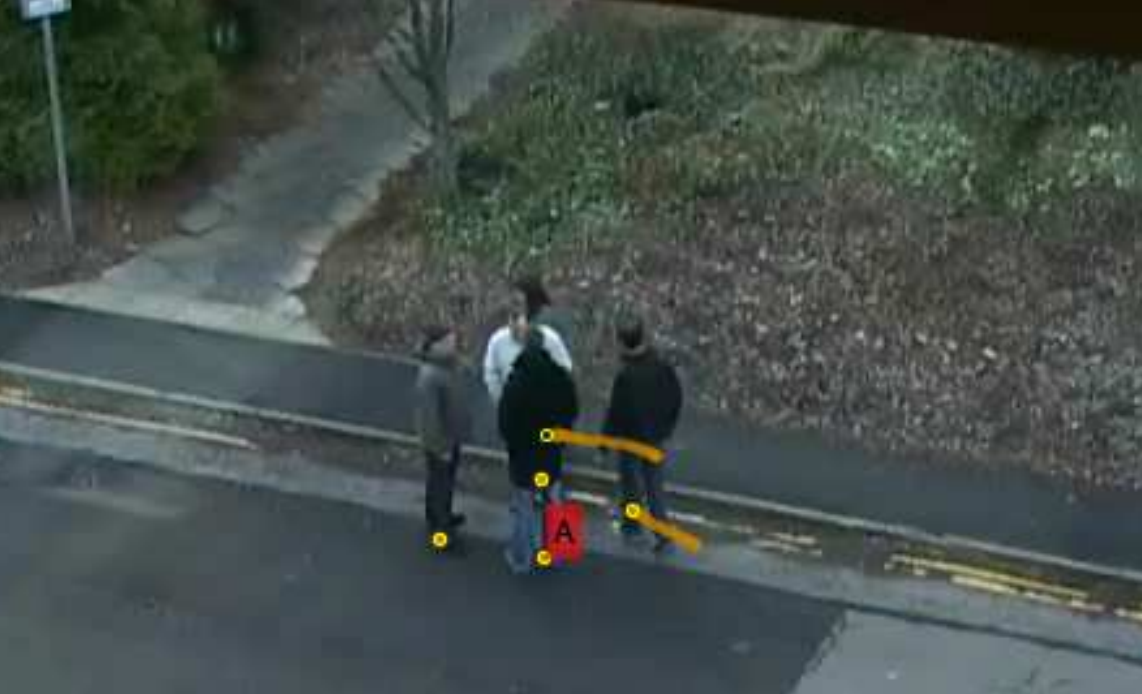}}~
\vspace{-0.25cm}
	\subfloat[\label{fig:split}Splitting]{\includegraphics[trim={3.5cm 5.5cm 1.8cm 0.5cm},clip, scale=0.4]{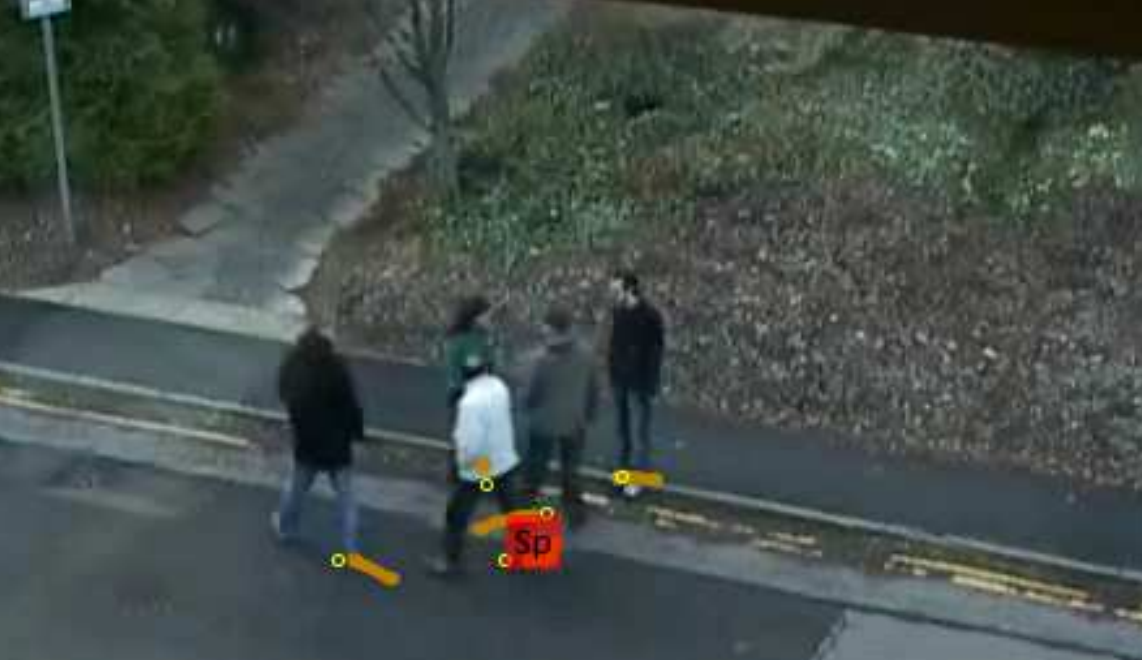}}~
	
     \label{types}
\caption{\protect\subref{fig:crowd2} and \protect\subref{fig:crowd1} give examples of structured and unstructured crowd. Output of the proposed algorithm: \protect\subref{fig:stand} - \protect\subref{fig:split} show groups with different activities: Standing (\textbf{St}), Walking (\textbf{W}), Splitting (\textbf{Sp}) and Approaching (\textbf{A}). Tracklets for some of the agents over past few frames are also shown. Each color represents a detected group (Best viewed in color). The videos are from \textit{BEHAVE}~\cite{behave} and \textit{CUHK}~\cite{scene} datasets.}
\end{figure*}

We believe that a moderately dense crowd consists of various groups which form a primary entity of a crowd \cite{scene, vision} whereas a highly dense crowd can be considered to form a single group and a highly sparse crowd might have groups with cardinality of one. Together, they guide the overall behavior of the crowd and individually influence the actions of the members. Therefore, group level analysis and hence group detection becomes important in crowd analysis. We define a group as a set of individuals (agents) having some sort of interactions \eg the group members are walking together. Spatial proximity is also necessary to form a group; if there are agents with a similar motion pattern but are far away from each other, they do not form a group as per our definition. Each group has its own set of goals that leads to various interaction patterns among the members of the group. The collective behavior of these constituent groups identifies the global crowd behavior which can vary from a highly structured to a completely unstructured pattern. In case of a structured crowd, for example $-$ marching of soldiers, all groups are in coordination and share the same goal (see Fig.\ref{fig:crowd2}); whereas in an unstructured crowd, for example $-$ at railway station or at a shopping complex, there are multiple groups with different goals (see Fig.\ref{fig:crowd1}). We are interested in understanding these different types of crowd behaviors at various levels by exploiting motion information of individuals. The paper makes the following contributions:

\begin{enumerate}

\item{} A framework is proposed to model the collective motion of the crowd by a first order dynamical system. The model captures the interaction patterns among the individuals. Although, the proposed model does not capture any possible non-linear relations, its usefulness for short-term analysis has been verified experimentally. We also provide an optimization formulation for the estimation of the interaction matrix under the constraints of spatial proximity, temporal continuity and sparsity of inter-agent relationship.

\item{}Since the interaction matrix is learned from the trajectory data, it captures the spatio-temporal patterns present among the agents. We observe that the eigenvectors of the interaction matrix reflect the spatio-temporal patterns. Thus, we propose a spectral clustering \cite{spectral} based algorithm to identify the groups present in the scene. Extensive experimentation on various datasets demonstrates the effectiveness of the algorithm.

\item{} We also demonstrate how the activities can be classified at three different levels $-$ at atomic (individual) level, at group level and at crowd level. The eigenvalues of the interaction matrix characterize various group and individual activities $-$ Fig \ref{fig:stand}-\ref{fig:split} show examples of activities at group level. At crowd level, we employ group level features to identify the behavior of the crowd. We classify the crowd videos in one of the 8 categories as defined by \cite{scene} and demonstrate its performance in terms of classification accuracy.

\end{enumerate}

The remaining part of  the paper is organized as follows. Next section reviews the related literature. Section \ref{model} explains the proposed mathematical formulation followed by group detection algorithm in Section \ref{group}. Detection of group activity and atomic activity is discussed in Section \ref{act}. We look at crowd video classification in Section \ref{class}. The experimental results are presented in Section \ref{exp} followed by conclusions in Section \ref{conclusion}. 

\section{Related Work}
There are numerous research papers in the challenging and interesting area of crowd behavior analysis. There are several holistic approaches (\eg~\cite{mehran},~\cite{solmaz},~\cite{motion},~\cite{dc},~\cite{anamoly2}) as well as particle based algorithms (\eg~\cite{sfm1},~\cite{sethi},~\cite{vision},~\cite{cf},~\cite{scs}) in the literature. Holistic methods analyze crowd as a single entity and ignore individuals or groups. In many papers, a dense crowd is considered analogous to fluid and hence concepts from fluid mechanics are applied for analysis. Mehran \etal in  \cite{mehran} present streakline representation of crowd flow for behavior analysis. Solmaz \textit{et al.} recognize crowd behaviors such as bottlenecks, fountainheads, lanes, arches and blocks through stability analysis of a dynamical system \cite{solmaz}. Benabbas \etal detect motion patterns and events in the crowded scenes by modeling motion and velocity at each spatial location \cite{motion}. In \cite{dc}, Lin \etal find a coherent motion regions in the video by generating thermal energy field.

The agent based approaches analyze each individual or group to discover the global behavior. Solera \etal propose correlation clustering based group detection which uses socially constrained features. Shao \etal introduce a collective transition prior in \cite{scene} and represent each group by a Markov chain. They define interesting group descriptors which proved to be useful in group state analysis and crowd classification. In \cite{sethi}, Sethi and Chowdhury propose a phase space algorithm to identify pairwise correlation between the motion patterns. Ge \etal find groups by hierarchical clustering based on pairwise velocities and distance \cite{Ge},~\cite{vision}. Zhou \etal find groups by using coherent filtering \cite{cf}. They propose a coherent neighbor invariance property which characterizes coherently moving individuals. Sochman \etal~\cite{soch} infer groups based on social force model \cite{sfm1}. They define a pairwise group activity confidence to identify groups. Srikrishnan and Chaudhuri in \cite{sri} define a linear cyclic pursuit based framework for collective motion modelling with the goal of short-term prediction. But they do not explore group detection and there is no analysis of crowd behavior. In the interesting work of \cite{scs}, they consider group detection as a clustering problem and learn a socially meaningful pairwise affinity under Structural SVM framework.  

Most of the particle based algorithms compute pairwise velocity and spatial cues to find the groups hierarchically. They do not model spatio-temporal patterns of the agents collectively which might capture more complex interactions. Additionally, most of the methods assume a constant velocity motion model which is not valid for many scenarios. To address these limitations in the paper,  we propose to model motion trajectories collectively instead of individually or pairwise. Also instead of relying on spatio-temporal information directly (which is prone to noise) for group detection, we use spectral clustering to identify groups.

\section{Mathematical Formulation}
\label{model}

We define a group as a set of agents having spatial proximity and some sort of interaction. In general, such interactions are complex and non-linear in nature. We approximate these interactions locally in time by a first order dynamical model. Note that we refer by agent an individual entity (represented by a point to be tracked) in the crowd.  

\subsection{Proposed Interaction Model}
We model the collective relationship among the agents by a first order affine system. Our hypothesis is based on the intuition that each agent takes into consideration ($a$) the movement of other agents present nearby and ($b$) her/his desired goal, while taking the next step. To capture these two intuitions, our model relates the next position of each agent to the current positions of all the agents including herself/himself. Let $\mathbf x(k) = [x_1(k), x_2(k), ..., x_N(k)]^T$, then


\begin{equation}
\mathbf x(k+1) =[\mathbf A | \mathbf a] 
\begin{bmatrix} \mathbf x(k) \\  1 \end{bmatrix}
= \mathbf A' \mathbf x'(k)
\label{eq:model}
\end{equation}\\

where  $N$ is the total number of agents, $\mathbf A \in \mathbb R^{N \times N}$, $\mathbf A' \in \mathbb R^{N \times (N+1)}$, $\mathbf a \in \mathbb R^{N \times 1}$ is the bias, $\mathbf x'(k) \in \mathbb R^{(N+1) \times1}$ and $x_i(k) \in \mathbb R$  is the location of the $i^{th}$ agent at time instant $k$ along the $x$-axis. We call $\mathbf A$ as the interaction matrix which captures the evolution of an agent as a function of all agents present in the scene. Note that $\mathbf A$ has no assumption on its form and entries. It need not be symmetric i.e. agent $i$ may not depend on agent $j$ in the same way as agent $j$ depends on agent. For example, consider a case where agent $i$ is stationary and agent $j$ approaches him/her. Since their behaviors are not symmetric with respect to each other, we assume that it implies $a_{ij}\neq a_{ji}$. 

In this paper, it is assumed that the motions along $x$ and $y$ directions are independent and hence can be analyzed independently. The corresponding model along $y$ direction is $\mathbf y(k+1)=\mathbf B \mathbf y(k)+\mathbf b$. In the rest of the paper, we discuss the solution for matrix $\mathbf A$ noting this fact that the same process is also carried out for $\mathbf B$. In the end, the outputs from both the models are combined appropriately to get the final output. We expect matrices $\mathbf A$ and $\mathbf B$ to be dependent on crowd motion. Since crowd behavior might change with time, the interaction matrix is time varying in nature, which we represent as $\mathbf A_k$ where $k$ is a time instant. Assuming $\mathbf A$ has $N$ independent eigenvectors, the general solution to Eq.(\ref{eq:model}) is given as

\begin{equation}
\mathbf x(k) = \sum_{\substack{i=1\\ \lambda_i \neq 1}}^{N}\{c_i\lambda_i^k\mathbf e_i + d_i\frac{(\lambda_i^k-1)}{\lambda_i-1}\mathbf e_i\}+\sum_{\substack{i=1\\ \lambda_i = 1}}^{N}(c_i+kd_i)\mathbf e_i,
\label{eq:sol}
\end{equation}

where $\lambda_i$ is the $i^{th}$ eigenvalue, $\mathbf e_i$ is the corresponding normalized eigenvector,  $c_i$ and $d_i$ are the corresponding constant coefficients that depend on the initial condition and $\mathbf a$ respectively. Different values of $\lambda_i$ and $\mathbf e_i$ generate various motion patterns for an agent. These patterns can be associated to different motion tracks generated by an agent while walking, approaching, splitting or being stationary. For example, an agent is stationary if $\lambda_1=1$ and $d_1=0$ at location $c_1\mathbf e_1$ or an agent is moving with a constant speed if $\lambda_1=1$ and $d_1 \neq 0$. Hence, this more generalized model is appropriate for modeling temporally localized complex motions.    

\begin{figure*}
\centering
	\subfloat[\label{fig:pred_err}]{\includegraphics[height=3.8cm, width=0.4\textwidth]{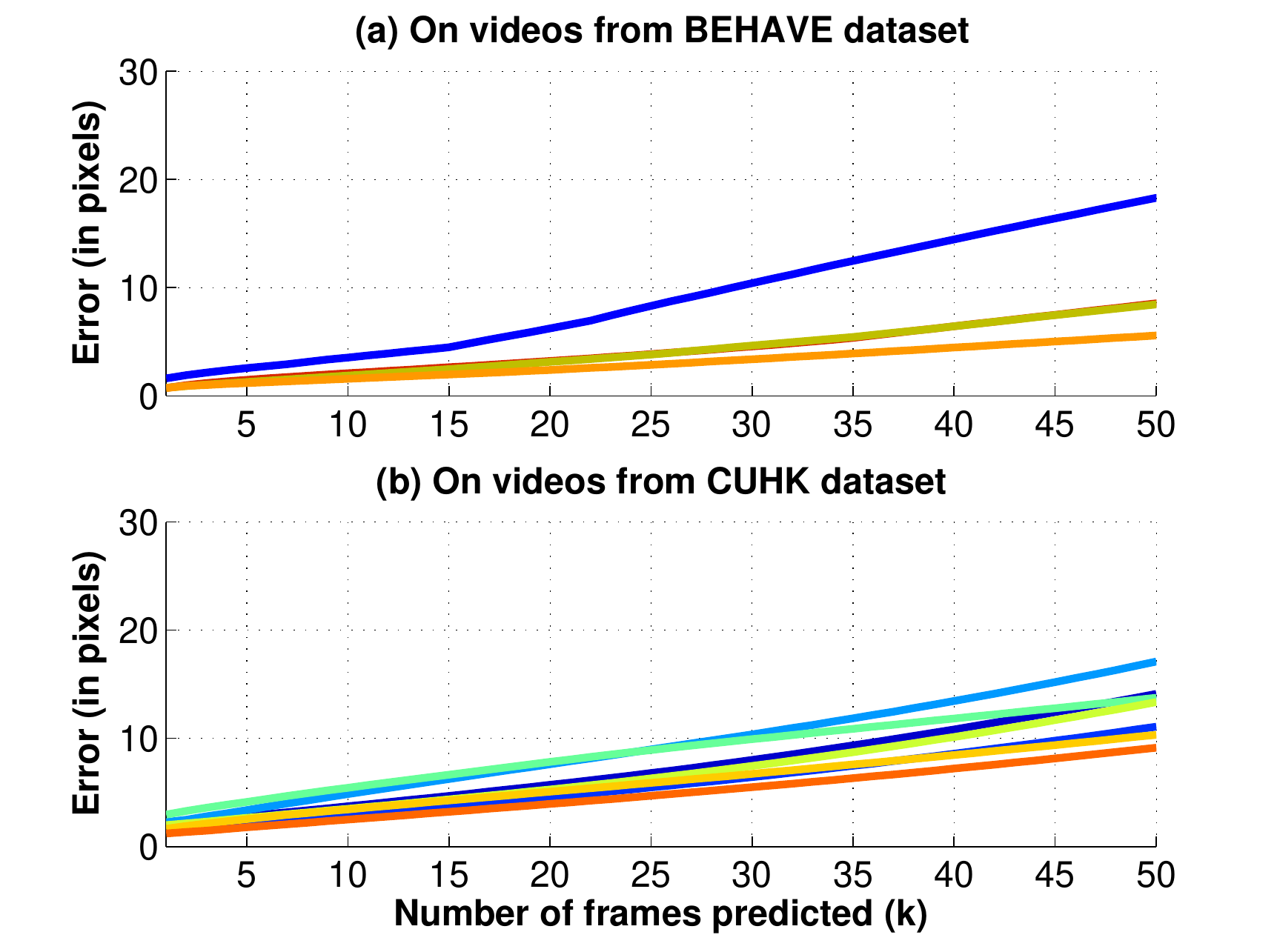}}~~~~~~
	\subfloat[\label{fig:ng1}]{\includegraphics[scale=0.37]{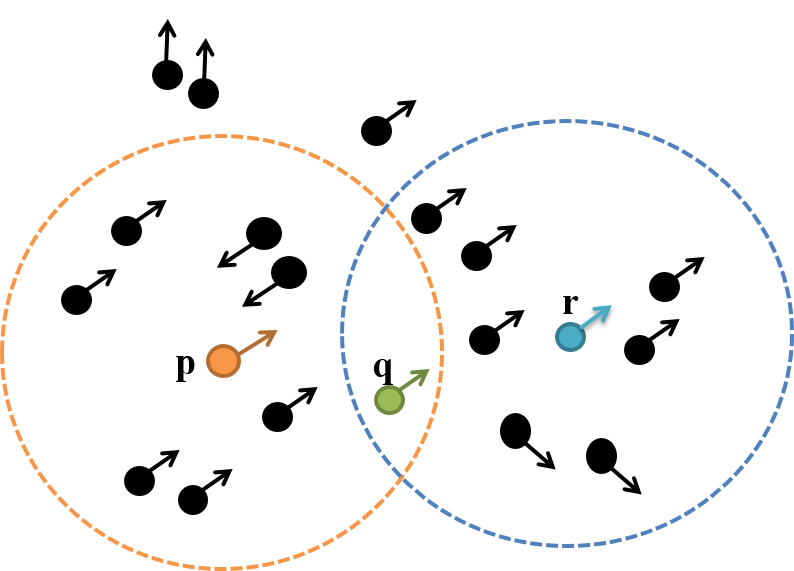}}
\caption{\protect\subref{fig:pred_err}  \textbf{Illustration of suitability of the proposed model}: Average $k$-step prediction error for sample videos from \textit{BEHAVE} and \textit{CUHK} datasets, each curve corresponds to a different video. \protect\subref{fig:ng1} \textbf{Neighborhood criteria}: Spatial neighborhoods around agents \textbf p and \textbf r are represented as circles around them. There are a total of 20 agents in the scene out of which only 8 are neighbors of \textbf p. Estimation of elements of a row of $\mathbf A$ corresponding to agent \textbf p, considering all agents present in the scene requires $2.5\times 20=50$ previous video frames (assuming $L=2.5N$). While the use of neighborhood constraint reduces this to $2.5*9\approx 23$ frames.}
\end{figure*}

\subsection{Validation of the Model}
We use an average $k$-step prediction error as a measure to test the validity of the proposed model on real videos. Fig.~\ref{fig:pred_err} shows average errors for different step size prediction on videos from \textit{BEHAVE} and \textit{CUHK} datasets, each curve corresponding to a different video. The $k$-step prediction error at any time instant $n$ is calculated as follows:

\begin{equation}
E_n(k)=\frac{1}{kN}\sum_{i=1}^{k}\sum_{j=1}^N |x_j^{actual}(n+i)-x_j^{pred}(n+i)| ,
\label{eq:pr}
\end{equation}

It may be noted that matrix $\mathbf A$ is estimated from the latest video frames up to $n$ and then Eq.~\ref{eq:model} is used to obtain $x_j^{pred}$. The $k$-step prediction error for the video is obtained by averaging $E_n(k)$ over all the frames of the video. As expected, error increases with $k$ but with a marginal increment. We observe that, for both the databases, prediction is quite valid up to 1-1.5 seconds (about 30 frames). Since the model assumes that the interaction remains same over $L$ frames, Fig~\ref{fig:pred_err} suggests that one can select $L$ upto 30 frames without introducing much error. These error plots show that the proposed model is suitable for short-term analysis, which is the underlying theme of the proposed algorithm. 

\subsection{Estimation of Interaction Model Parameters}
The matrix $\mathbf A$ and vector $\mathbf a$ at any time instant are learned from the immediate past trajectory data of all the agents in a least squares framework. We update $\mathbf{A}$ and $\mathbf a$ with each incoming frame as interaction patterns may change over the time. In addition, sudden changes in these interactions are unlikely. Therefore it is desired that the entries of $\mathbf A$ and $\mathbf a$ do not change drastically in consecutive time instants $-$ we assume them to be varying smoothly over time. We incorporate this constraint by minimizing \textit{$l_2$} norm of the difference between the current matrix $\mathbf A'_k$ and the previous estimate at $(k-1)^{th}$ instant. Furthermore for crowded scenes, it is unlikely that an agent's motion depends on all the agents present in the neighborhood. We capture this sparse relationship in $\mathbf A'_k$ by minimizing \textit{$l_1$} norm of $\mathbf A'_k$. 

Adding these constraints to the cost function, the final formulation at $k^{th}$ time instant becomes:

\begin{eqnarray}
\mathbf {\hat{A}}_k' &=&\arg\min\limits_{{{\mathbf A'_k \in\mathbf{R}^{N \times (N+1)}}}}\Big\{||\mathbf A'_k\mathbf X_{k-L}^{k-1}-\mathbf X_{k-L+1}^k||_2^2  \nonumber \\
&+& r_1||\mathbf A'_k-\mathbf A'_{k-1}||_2^2+r_2||\mathbf{A'_k}||_1\Big\},
\label{eq:form}
\end{eqnarray}
where $\mathbf X_i^j \in \mathbb{R}^{N+1 \times L}$ contains the positions of all $N$ agents from ~$i^{th}$ to $j^{th}$ frames concatenated together with an appended row of ones to account for the bias, $\mathbf A'_{k-1}$ is the estimate at the previous frame and $r_1$ and $r_2$ are appropriate regularization parameters. Note that we will use $\mathbf A'$ instead of $\mathbf A'_k$ for notational convenience. 

One requires at least $L \geq (N+1)$ past positions to solve the Eq.~\ref{eq:form}. Therefore the interaction pattern is assumed to remain constant over $L$ frames. Hence we want $L$ to be small enough to capture the short-term linear relationship among the agents. A large $N$ (in crowded scenes) leads to two major problems: ($i$) longer trajectories (i.e. higher $L$) are required to learn the interaction matrix $\mathbf A'$ as $L\geq N+1$ which may not be available and ($ii$) the interaction may not remain constant over past $L$ positions for high values of $L$ as discussed before and we would like to keep $L\leq 30$ as discussed in the previous section. To address these problems, we identify spatial neighbors of each agent separately and learn only the corresponding entries in the matrix (one row at a time). The neighborhood is defined as follows $-$ the agent \textbf p is a neighbor to the agent \textbf q if $dist(\textbf p,\textbf q)<R_{\textbf p}$. The value of $R_{\textbf p}$ is decided so as to satisfy the constraint $L\leq30$. The intuition for enforcing the neighborhood criteria is that it is unlikely that far away agents influence the motion of an agent. The advantage is that the shorter trajectories are now sufficient as the number of entries of $\mathbf A'$ to be learned are lesser. Note that we estimate matrix $\mathbf A'$ in a row-wise manner where the $i^{th}$ row has number of entries to be estimated as equal to one more than the number of the neighbors of agent $i$. Further, there could be an agent within the spatial proximity of another agent but there may not be any interaction between them. Hence it is required that the corresponding entry in the matrix $\mathbf A'$ should be zero. This is enforced by adding sparsity constraint in Eq.~\ref{eq:form}. We use L1General package developed by Schmidt~\cite{L1G} for solving L1-regularization problems.

For an illustration, see Fig.\ref{fig:ng1}. There are a total of $N=20$ agents present in the scene. Estimation of the row of matrix $\mathbf A$ corresponding to agent \textbf p requires 50 previous frames (assuming $L=2.5N$) whereas the neighborhood based estimation reduces this to 23. Also consider a case where agents \textbf p and \textbf r interact with each other but are not within the spatial proximity owing to neighborhood constraint. The interaction is captured when intersection of neighborhoods of \textbf p and \textbf r has at least one interacting agent, in this case its \textbf q who is in the spatial proximity of both. 

\section{Group Detection}
\label{group} 

In this section, we discuss an algorithm for identifying the groups present in the scene. As seen from Eq.~\ref{eq:sol}, the general solution is a linear combination of eigenvectors at any time instant $k$. Notice that if the corresponding entries of any two rows of the eigenvector matrix are similar, the corresponding agents form a group. This group information is not available from the position vector alone at a particular time instant $\mathbf x(k)$ because temporal evolution is also an important factor in deciding the groups.  Since the eigenvectors are learned from the trajectories collectively, it encapsulates spatio-temporal evolution of the agents and hence can be exploited for group detection. 

Let the eigenvector matrix contain all the eigenvectors column-wise. We define a mapping for $i^{th}$ agent as ${f(x_i):x_i\in\mathbb R}\rightarrow\mathbf z_i=(e_{i1}, e_{i2}, \ldots, e_{ir})^T\in\mathbb R^{r \times 1}$ where $e_{ji}$ is the $i^{th}$ entry of $j^{th}$ eigenvector of interaction matrix $\mathbf A$ and $r$ is the number of significant eigenvalues. A clustering algorithm is applied on the points \{$\mathbf z_i\}, \forall i = 1, 2,\ldots, N$ to identify the groups. The clustering algorithm runs on the components of eigenvectors, therefore this algorithm falls in the category of spectral clustering~\cite{spectral}. Since the number of groups is unknown, we apply a threshold based clustering. The adaptive threshold used for the $i^{th}$ point is $c|\mathbf z_i|$, where $|\mathbf z_i|$ is its magnitude and $c$ is found empirically. For example, all the agents within the distance of $c|\mathbf z_1|$ from $\mathbf z_1$ will form a group with agent $1$. In this way, all the groups are obtained. We consider only significant eigenvectors (with $ |\lambda| \geq 0.90$) of $\mathbf A$ for group detection since the response from the eigenvectors with $|\lambda|<0.9$ dies down to an insignificant level within the period of $L$ frames. 

It may be noted that this group detection algorithm remains the same in the case where $\mathbf A$ does not have $N$ independent eigenvectors. In such a case, the clustering algorithm runs on generalized eigenvectors. 

\section{Group Activity Identification}
\label{act}
While the eigenvectors identify the groups, the eigenvalues can be used to determine the activity of a group. We employ the same  model mentioned in Eq.~\ref{eq:model} for the group \textit{g} to estimate its interaction matrix $\mathbf A^ \textit{g}$ and $\mathbf a^ \textit{g}$. We do not use the submatrix formed by the agents of the group \textit{g} in the previously learned matrix $\mathbf A' = [\mathbf A | \mathbf a]$ to get $\mathbf A^{ \textit{g}'}=[\mathbf {A}^ \textit{g} | \mathbf a^{\textit{g}}]$. This is to get a refined matrix for the group and avoid any possible interference from the outside agents in the estimation.  Let $\mathbf x^ \textit{g}(k)=[x^ \textit{g}_1(k), x^ \textit{g}_2(k),\ldots, x^ \textit{g}_M(k)]^T$, where $M$ is the cardinality of the group  \textit{g} and $x^ \textit{g}_i(k)$ is the position of the $i^{th}$ agent of the group at time instant $k$. To learn matrix $\mathbf A^ \textit{g'}$ at $k^{th}$ time instant, we define a similar optimization framework as follows, where the second term enforces temporal continuity in the activity but unlike Eq.~\ref{eq:form}, there is no need for sparsity constraint as, by definition, all agents in a group interact. Therefore,

\begin{eqnarray}
\mathbf {\hat{A}}_k^{ \textit{g'}}&=&\arg\min\limits_{{{\mathbf A^ \textit{g'}_k \in\mathbf{R}^{M \times (M+1)}}}}\Big\{||\mathbf A^ \textit{g'}_k\mathbf X_{k-L}^{k-1}-\mathbf X_{k-L+1}^k||_2^2 \nonumber \\
&+&\lambda||\mathbf A^ \textit{g'}_k-\mathbf A^ \textit{g'}_{k-1}||_2^2\Big\}
\end{eqnarray}

\begin{figure*}
\centering
	\subfloat{\includegraphics[scale=0.5]{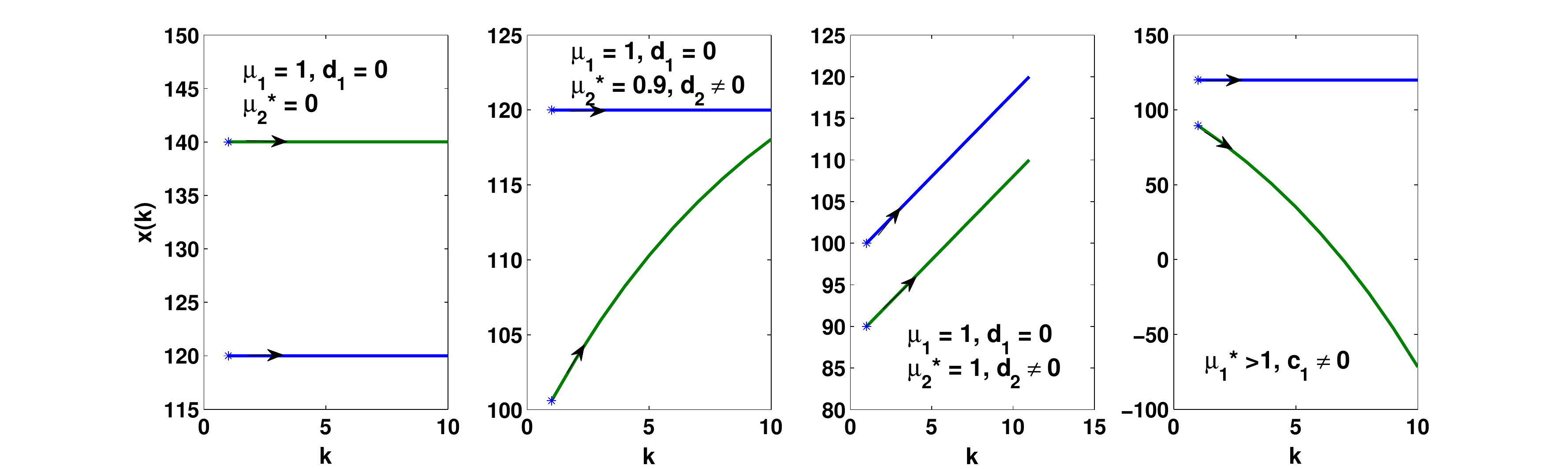}}~ 
	\caption{Illustration of group activity - \textit{Stationary}, \textit{Approaching}, \textit{Walking} and \textit{Splitting} respectively from the estimated model parameters for a group consisting of two members. Eigenvalue with $*$ is the activity deciding eigenvalue. See the text for details. }
\label{fig:syn}
\end{figure*}

Assuming $\mathbf A^ \textit{g}$ to be again diagonalizable, the general solution is similar as given in Eq. \ref{eq:sol}. The velocity vector $\mathbf v(k)$ for the group \textit{g} can be written as
\begin{equation}
\mathbf v(k) = \sum_{i=1}^{M}\{c_i(\mu_i-1)\mu_i^{k-1} + d_i\mu_i^{k-1}\}\mathbf u_i,
\label{eq:vel_sol}
\end{equation} 

where $|\mu_1|\geq|\mu_2| \ldots \geq|\mu_M|$ are the eigenvalues of $\mathbf A^{\textit{g}}$. Since some of the coefficients $c_i$ and $d_i$ could be zero, let $\mu_j$ be the largest eigenvalue for which at least one of the coefficients $c_i$ or $d_i$ is non-zero. Now we state how different values of $\mu_j$ characterize various activities: 

\begin{enumerate}
\item \textbf{Stationary}: A group is stationary if $|\mu_j|=0$ indicating all the eigenvalues (with at least one non-zero coefficient) to be zero. That corresponds to zero velocity vector and hence the agents are stationary.  In the illustration shown in Fig.~\ref{fig:syn}(a), the deciding eigenvalue is $\mu_2$ which is 0. The two agents are stationary at locations 140 and 120 respectively.
\item \textbf{Approaching}: A group has an approaching members if $|\mu_j|<1$ as $\lim_{k\to\infty}\mathbf v(k)\to 0$. In the example shown in Fig.~\ref{fig:syn}(b), $j=2$. One agent is stationary at 120 while the other agent starts from the location 100 and approaches to the first one.
\item \textbf{Walking}: If $|\mu_j|=1$ then the group is walking with a constant velocity of $d_j\mathbf u_j$. In Fig.~\ref{fig:syn}(c), both the agents walk together and deciding eigenvalue corresponds to $j=2$. Note that we do not discriminate between walking and running in this work.  
\item \textbf{Splitting}: A group has a tendency for divergence if $|\mu_j|>1$ as $\lim_{k\to\infty}\mathbf v(k) \to \infty$.  In Fig.~\ref{fig:syn}(d), this corresponds to $j=1$. Initially the two agents were standing together and then the second agent starts moving away from the first one leading to split of the group. 
\end{enumerate}

This group activity detection method is dependent on eigenvalues and hence sensitive to perturbations in the measurements. To address this, we define threshold bands for crucial values of eigenvalues. For example, if $0.995<\mu<1.005$, we consider $\mu$ to be $1$ and if $\mu<0.5$ then it is considered as 0. 

\subsection{Atomic Activity Detection}
This algorithm is now extended to identification of individual's activity as follows. Let $x(k)$ denotes position of an agent at time $k$, then 
\begin{equation}
x(k+1)=\mu x(k)+ b
\end{equation}

The velocity $v(k)$ is as follows:
\begin{equation}
v(k)=(1-\mu)\mu^{k-1}x(0)+\mu^{k-1}b
\end{equation}

Note that there is no longer a activity called splitting as one needs at least two agents to define it. We identify following activities based on the value of $\mu$:
\begin{enumerate}
\item{\textbf{Stationary}}: An agent is stationary if $|\mu|=0$ at the location given by $b$. It is also stationary when $|\mu|=1$ and $b=0$. 
\item{\textbf{Stopping}}: $0<|\mu|<1$ indicates that the agent is stopping soon.
\item{\textbf{Walking}}: An agent is walking if $|\mu|>1$. Further, an agent is walking with a constant velocity if $|\mu|=1$ and $b \neq 0$. 
\end{enumerate}

Note that the group detection and activity recognition algorithms run in $x$ and $y$ directions independently and results need to be combined together. For group detection, a group is formed only if it is formed in both the directions. For example, let $Z_x=[1,1,2,1]$ and $Z_y=[2,1,2,2]$ be the label vectors (indicating assigned group number for all the four agents) obtained along $x$ and $y$ directions respectively. It says that agents \{1,2,4\} form a group along $x$ direction while \{1,3,4\} form a group along $y$ axis. Combining both the labels will result in the final label vector as $Z=[1,2,3,1]$ \ie~out of 4 agents, 1 and 4 are grouped together while agents 2 and 3 are singleton groups. To identify the final group activity from the two separate group activity estimates along $x$ and $y$ directions, we merge the two decisions according to the following priority sequence $-$ \textit{Splitting}$>$\textit{Walking}$>$\textit{Approaching}$>$\textit{Stationary}. For example, if a group has \textit{splitting} and \textit{approaching} activities in $x$ and $y$ directions respectively, the final group activity is \textit{splitting}.  

\section{Crowd Video Classification}
\label{class}
Having group level information in hand, we can use them in identifying the overall crowd behavior. Ability to identify crowd behavior enables crowd management systems to design and manage public places effectively to ensure safety and smooth operation. The overall crowd behavior is determined by how each group behaves. Depending on the synchronization among the groups, the behavior of crowd varies from being structured to unstructured. In this section, we define group level features that are useful for crowd video classification. We classify crowd videos into 8 classes as defined by \cite{scene}. The dataset containing 474 video clips covers a variety of videos. The eight classes are as follows:
\begin{description}
\item[$C_1$]: Mixed crowd
\item[$C_2$]: Well organized crowd following mainstream: 
\item[$C_3$]: Not well organized crowd following any mainstream
\item[$C_4$]: Crowd merge
\item[$C_5$]: Crowd split
\item[$C_6$]: Crowd crossing in opposite directions
\item[$C_7$]: Intervened escalator traffic
\item[$C_8$]: Smooth escalator traffic
\end{description}    

We employ group level features that cover low-level details such as motion information to high-level information such as group activities. The features are described as follows:

\begin{enumerate}

\item \textbf{Group density ($GD$)}: It is the ratio of number of groups by the total number of agents in the scene. A low value of $GD$ indicates highly structured crowd. For example, $GD$ for a group of marching soldiers is small whereas a mixed crowd has a higher group density. 

\item \textbf{Histogram of ${\lambda_{max}}$}: The histogram has three bins which are ${\lambda}_{max}\geq 1$, ${\lambda}_{max}<1$ and ${\lambda}_{max}=0$, where ${\lambda}_{max}$ is the largest eigenvalue of the interaction matrix for a group ($\mu_j$ from the last section). The value at a particular bin is the number of groups in a scene having ${\lambda}_{max}$ as defined by that bin. Left skewed histogram i.e. towards $\lambda_{max}\geq 1$ indicates moving crowd whereas right skewed histogram suggests more or less stationary crowd.

\item \textbf{Histogram of motion direction}: The motion direction of each member of a group is calculated from its trajectory data and the mean direction is assigned to the group. This histogram has eight bins covering $0\degree$ to $360 \degree$ with a bin size of $45\degree$. The bin value is the number of groups falling in that particular bin. The uniform histogram indicates a mixed crowd whereas a skewed histogram indicates directionality in the crowd movement.


\end{enumerate}

Since the analysis  is conducted independently in $x$ and $y$ directions; we get two histograms for $\lambda_{max}$, leading to final feature vector of length $1+2\times 3+8=15$. We use random forest (RF) as a classifier \cite{RF}. It consists of a multitude of decision trees that are trained from randomly sampled subsets of training dataset (bootstrap aggregating). This bootstrapping increases the performance by reducing the variance of the classifier. Also the split at each node of a tree is decided by $m$ features selected randomly out of $n$ features where $m<<n$. We use RF to classify a crowd video by training it with the above mentioned features. The classification results are discussed in next section.

\section{Experiments and Results}
\label{exp}
In this section, we discuss the performance of the proposed algorithms for group detection, group activity recognition and crowd video classification. We have tested our algorithms on various publicly available datasets containing real videos. We first discuss these datasets  followed by performance evaluation of the proposed algorithms.

\subsection{Datasets}
We tested our algorithms on different videos from various datasets contributed by several researchers namely \textit{CUHK}~\cite{scene}, \textit{BEHAVE}~\cite{behave}, \textit{BIWI Walking Pedestrians}~\cite{biwi}, \textit{Crowds-By-Example} (\textit{CBE})~\cite{cbe} and \textit{Vittorio Emanuele II Gallery} (VEIIG)~\cite{veiig}. \textit{CUHK} dataset is a comprehensive crowd video dataset containing 474 video clips covering various crowd behaviors with varying crowd density. \textit{BEHAVE} dataset has video clips with low crowd density and covering various group activities. \textit{BIWI} dataset contains two low density crowd videos (namely \textit{eth} and \textit{hotel}). \textit{CBE} has a medium density crowd video (\textit{student003}) recorded outside a university. These datasets collectively cover a large variety of crowd videos.

\subsection{Group Detection}
We tested group detection algorithm on all the 474 videos from \textit{CUHK} dataset and 3 video clips (having duration of more than 10 minutes in total) from \textit{BEHAVE} dataset. In case of videos from \textit{CUHK} dataset, we restricted our algorithm to run only on those data that have sufficiently long tracks, since some of the clips are too short to accommodate for an analysis of a large number of agents. We compared the proposed algorithm with other methods on these selected agents. The ground truth for \textit{CUHK} dataset was obtained manually. 

\begin{table}[h]
\centering
\caption{Performance comparison of different group detection algorithms on \textit{CUHK} dataset.}
\begin{tabular}{ |c|c|c|c|}

\cline{1-4}
\multicolumn{1}{|c|}{} & {~~~~CF} \cite{cf}~~~  & {~~~~CT} \cite{scene}~~~ & \textit{~~~Proposed~~}\\ 

\hline
\textit{~NMI~} & 0.66 & 0.69 & \textbf{0.86}\\ 
\hline
\textit{~Purity~} & 0.71 & 0.72 &\textbf{0.90} \\ 
\hline
\textit{~RI~} & 0.67&0.69 & \textbf{0.85}\\ 
\hline
\end{tabular}
\label{table:comp}
\end{table}

\begin{table*}
\centering
\caption{Performance comparison of the proposed group detection with~\cite{scs}}.
\label{table:comp_scs}
\begin{tabular}{l|l|l|l|l|l|l|}
\cline{1-7}
\multicolumn{1}{|c}{} \multirow{2}{*}{}                        & \multicolumn{2}{|c|}{Baseline {~\cite{scs}}}                                             & \multicolumn{2}{c|}{{~\cite{scs}}}                                                      & \multicolumn{2}{c|}{Proposed}                                                       \\ \cline{2-7} 
     \multicolumn{1}{|c|}{}                                         & \multicolumn{1}{c|}{\textit{\textbf{P}}} & \multicolumn{1}{c|}{\textit{\textbf{R}}} & \multicolumn{1}{c|}{\textit{\textbf{P}}} & \multicolumn{1}{c|}{\textit{\textbf{R}}} & \multicolumn{1}{c|}{\textit{\textbf{P}}} & \multicolumn{1}{c|}{\textit{\textbf{R}}} \\ \hline
\multicolumn{1}{|l|}{\textit{BIWI eth}}       & $72.4\pm4.4$                              & $65.2\pm3.4$                                & $91.8\pm1.2$                                & $94.2\pm0.9$                                & $\mathbf{95.78}$                                    & $\mathbf{96.15}$                                    \\ \hline
\multicolumn{1}{|l|}{\textit{CBE student003}} & $59.9\pm2.9$                                & $53.5\pm6.8$                                & $81.7\pm0.2$                                & $82.5\pm0.2$                                & $77.58$                                   &$\mathbf{85.90}$                                     \\ \hline
\multicolumn{1}{|l|}{VEIIG}                   &$49.2\pm9.9$                                 &$34.4\pm6.7$        & $84.12\pm0.6$                                    & $84.11\pm0.5$                                    & $82.97$                                    & $\mathbf{84.70}$                                    \\ \hline
\end{tabular}
\end{table*}

We compare the proposed algorithm for group detection with state-of-the-art methods by Shao \etal \cite{scene} and Zhou \etal \cite{cf}. For quantitative analysis on \textit{CUHK} videos, we randomly select two time instants for each video where we compare the proposed algorithm with other methods and the ground truth instead of manually deciding on the instants when the performance has to be evaluated. We use Normalized Mutual Information (NMI) \cite{nmi}, Purity \cite{purity} and Rand Index (RI) \cite{ri} which are widely used for evaluation of clustering algorithms. Table~\ref{table:comp} shows the comparison on these measures. It is quite evident from the table that the performance of the proposed algorithm far surpasses those of~\cite{scene} and~\cite{cf}. 

\begin{figure*}
\vspace{0.25cm}
\centering
\subfloat{\includegraphics[scale=0.18]{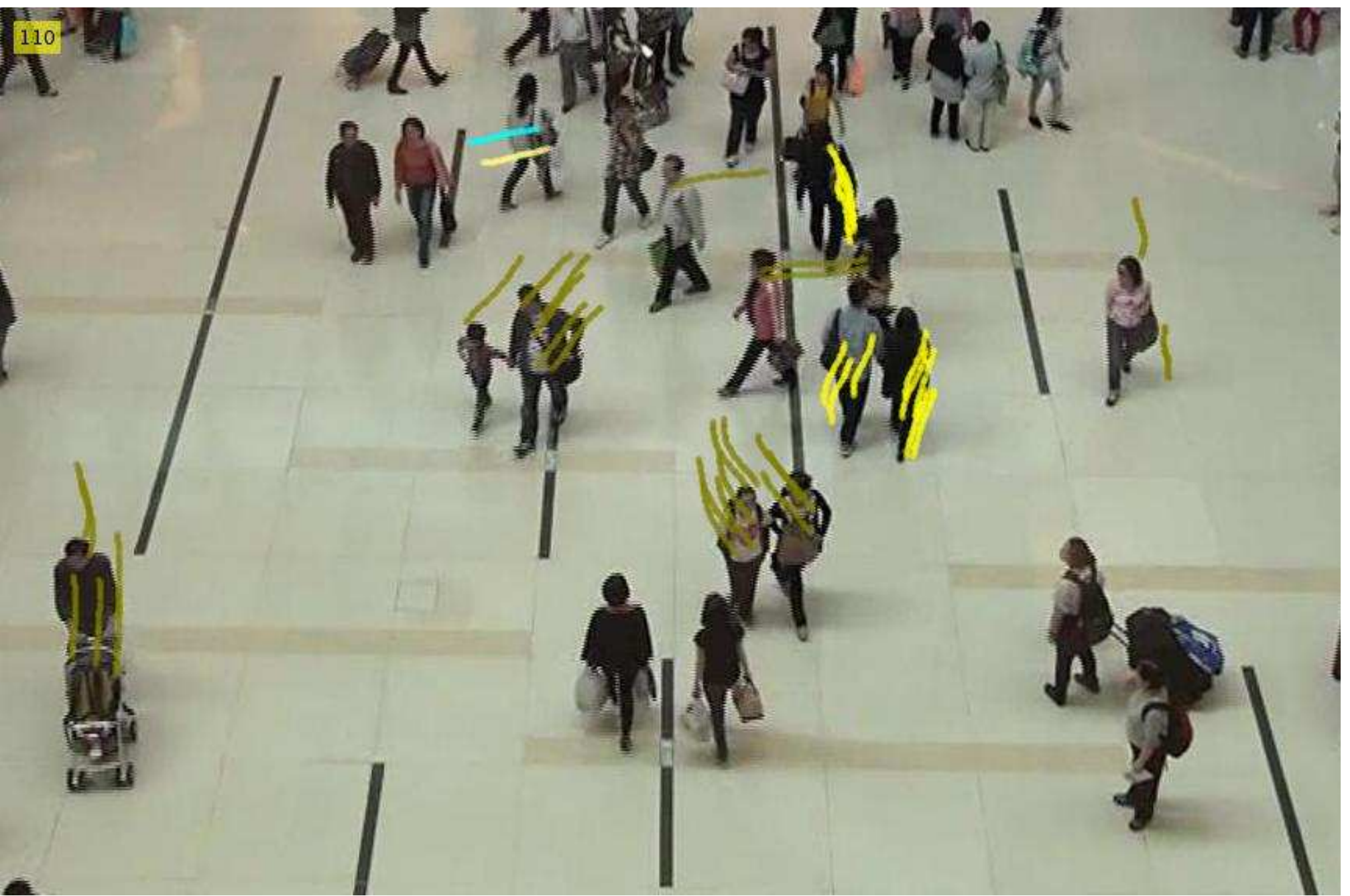}}\hspace{0.00001cm}
\clearsubcaptcounter
\subfloat{\includegraphics[scale=0.18]{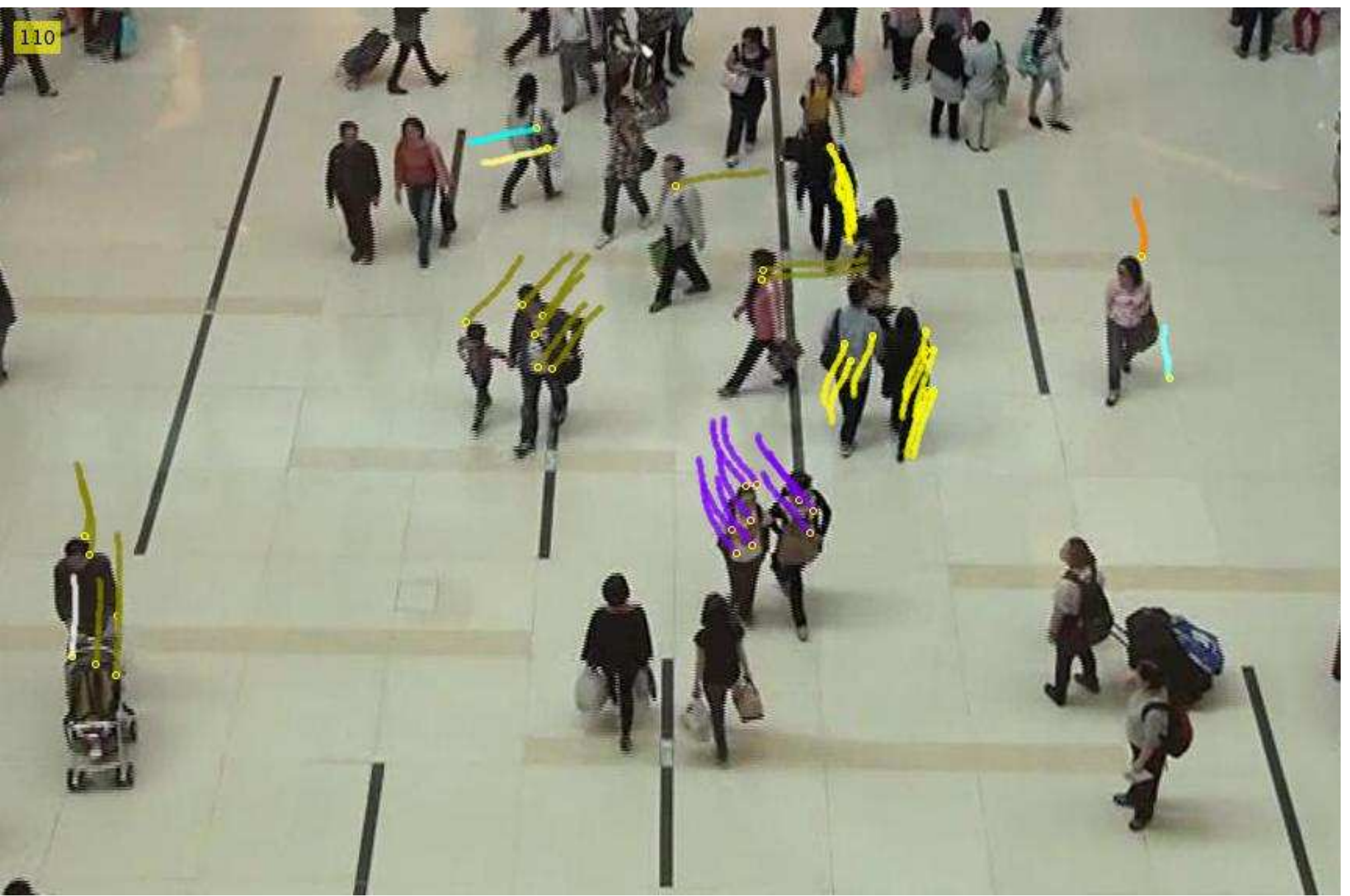}}\hspace{0.00001cm}
\clearsubcaptcounter
\subfloat{\includegraphics[scale=0.18]{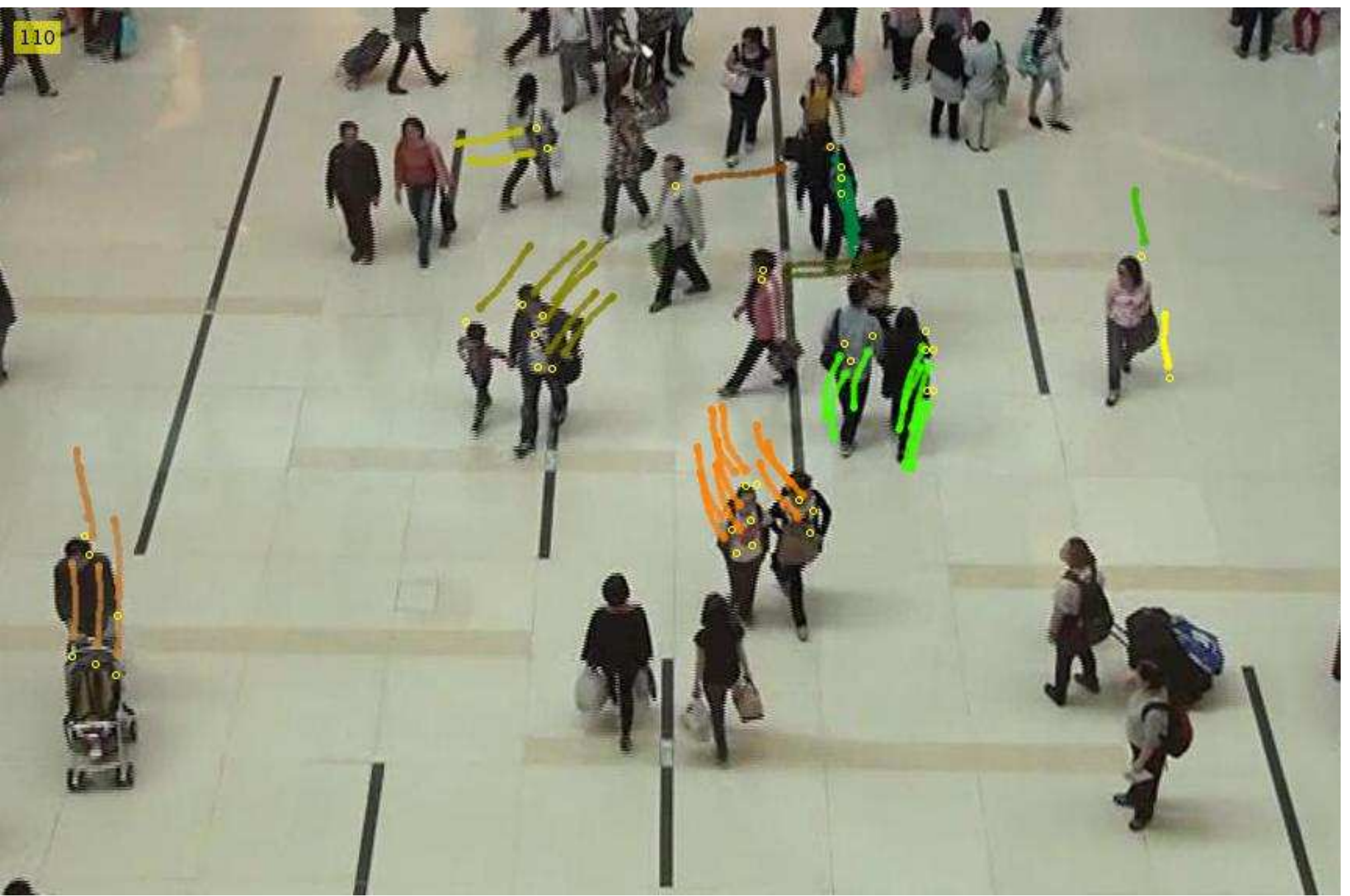}}\hspace{0.00001cm}
\clearsubcaptcounter
\subfloat{\includegraphics[scale=0.18]{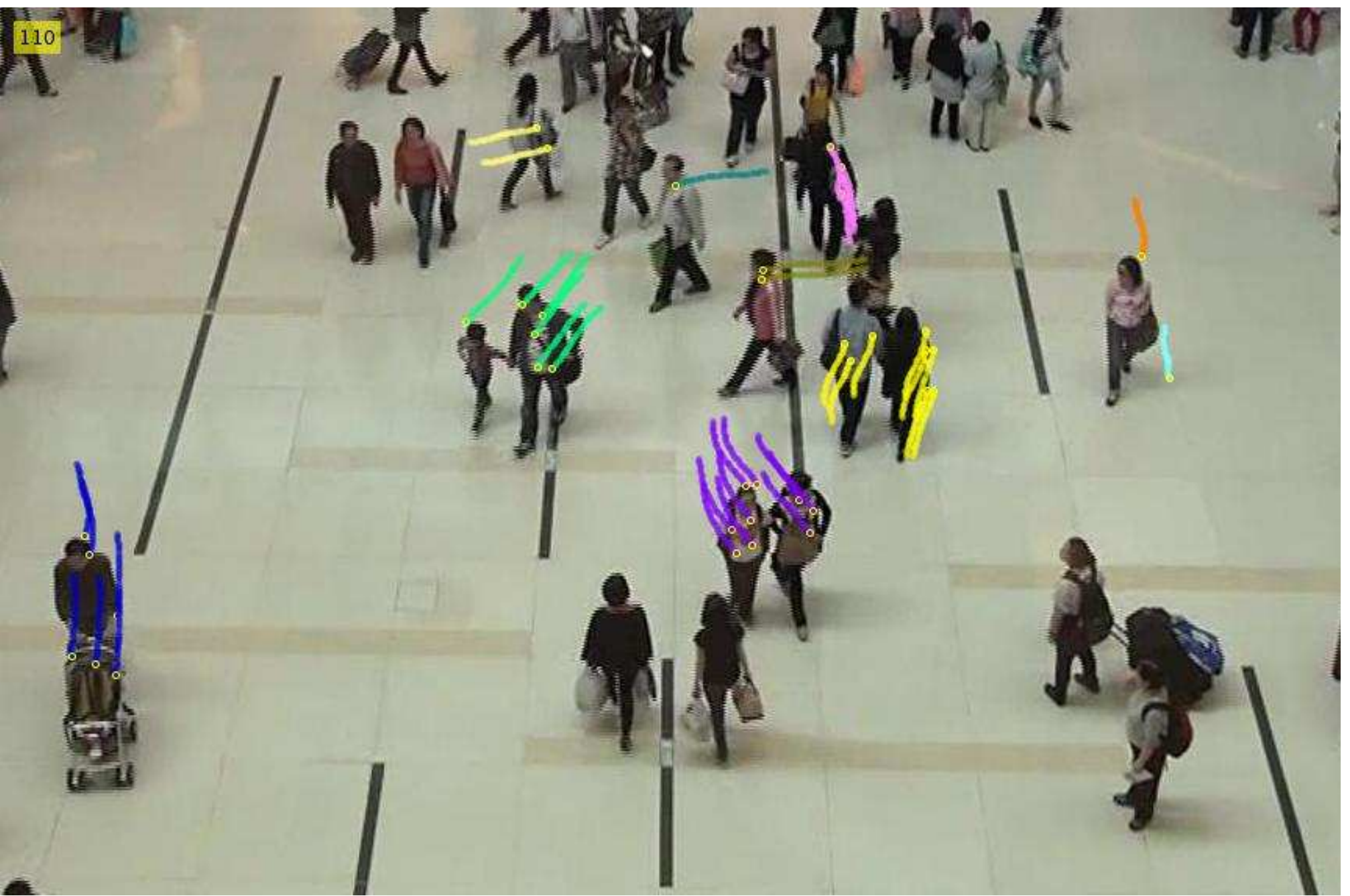}} \\
\clearsubcaptcounter
\vspace{-0.25cm}
	\subfloat{\includegraphics[scale=0.18]{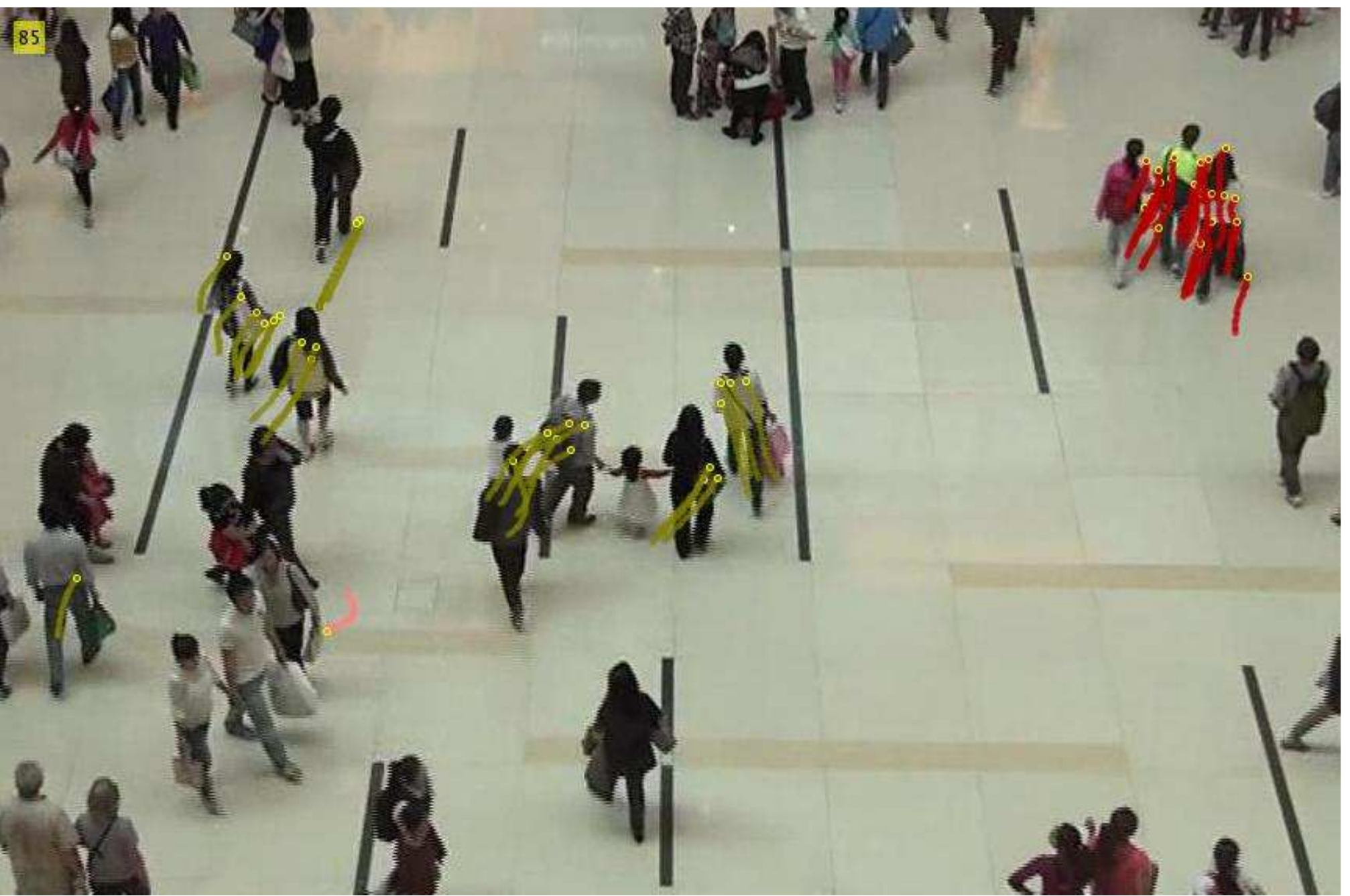}}\hspace{0.00001cm}
	\clearsubcaptcounter
	\subfloat{\includegraphics[scale=0.18]{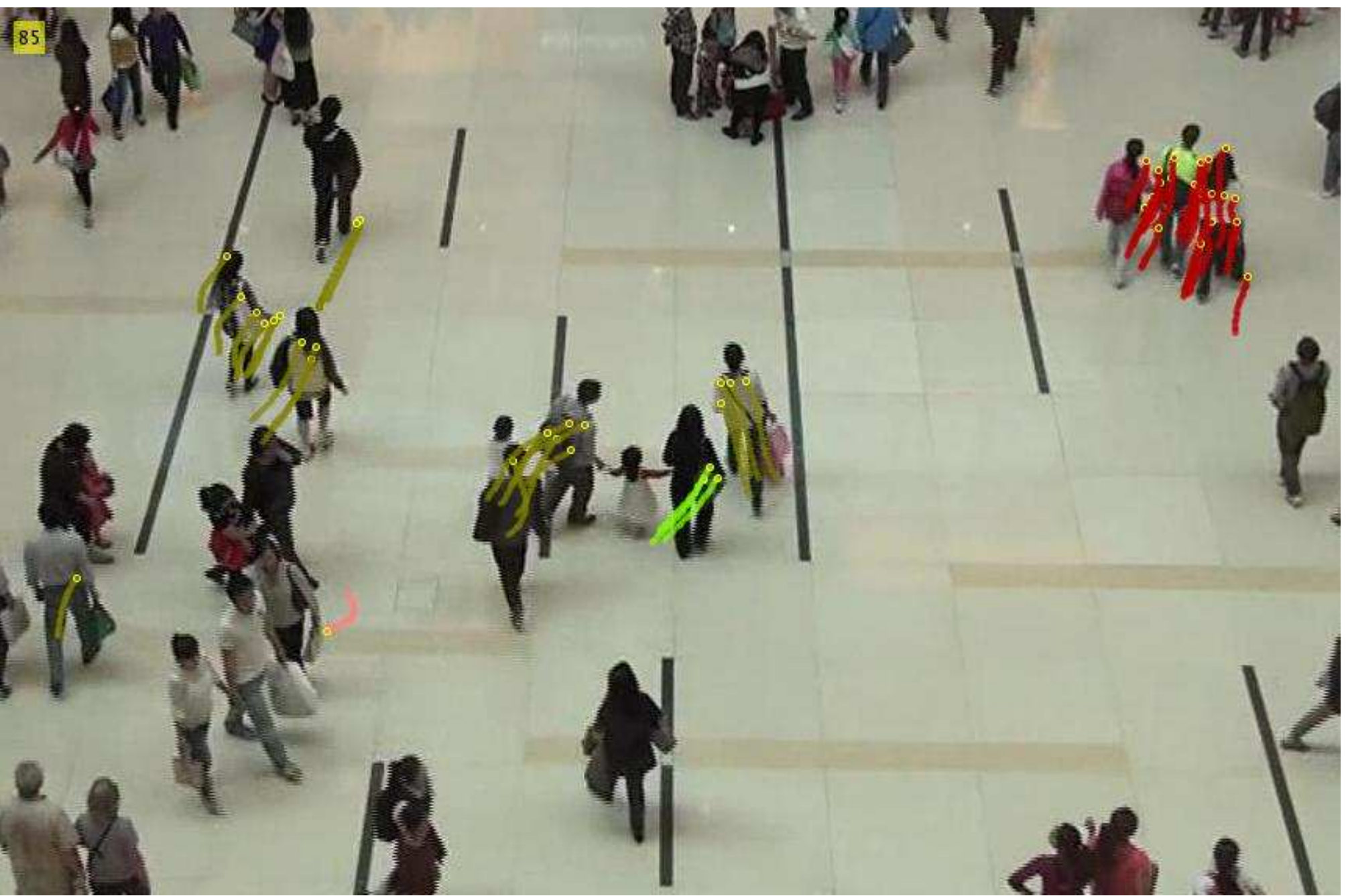}}\hspace{0.00001cm}
	\clearsubcaptcounter
	\subfloat{\includegraphics[scale=0.18]{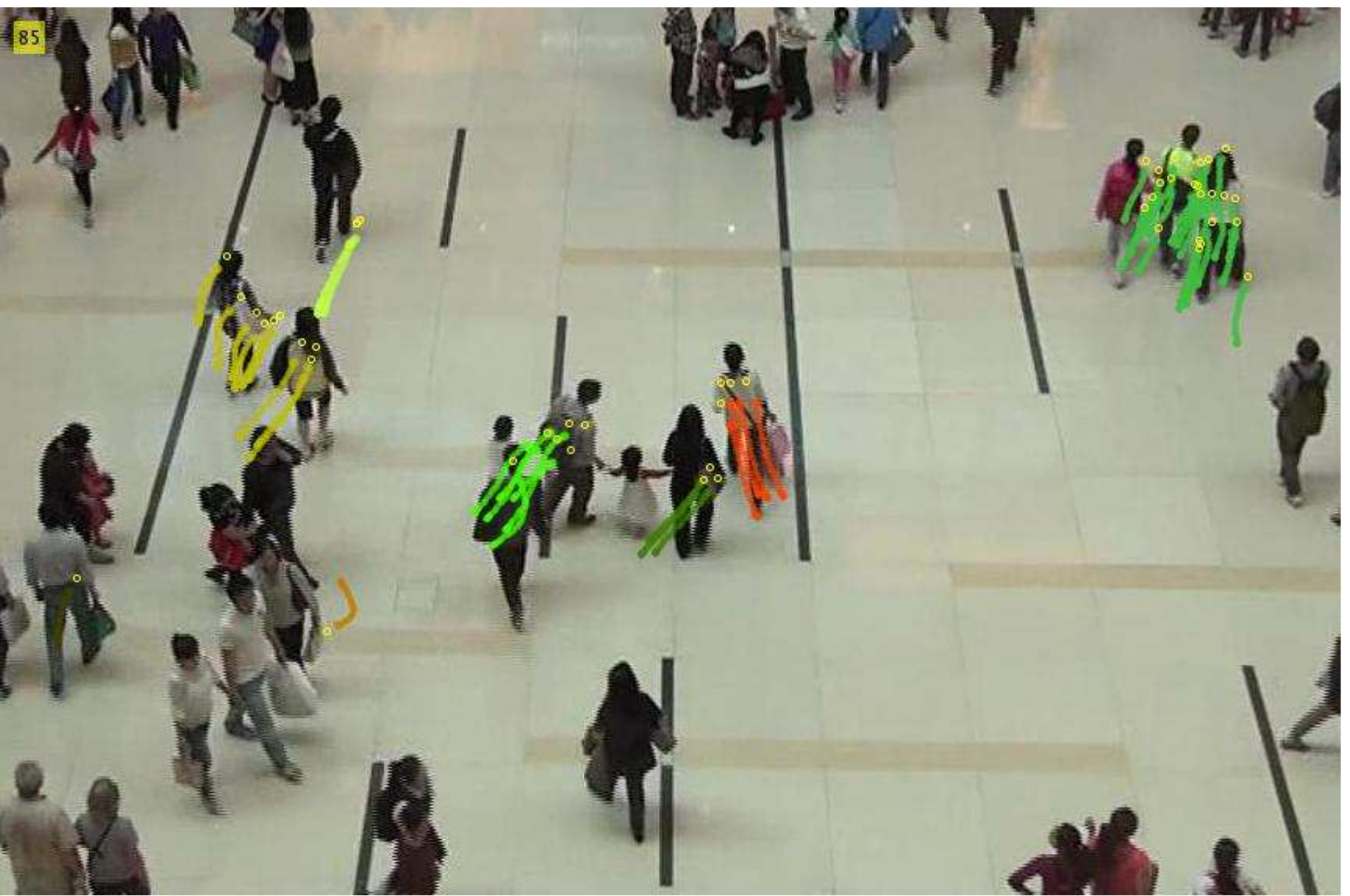}}\hspace{0.00001cm}
	\clearsubcaptcounter
	\subfloat{\includegraphics[scale=0.18]{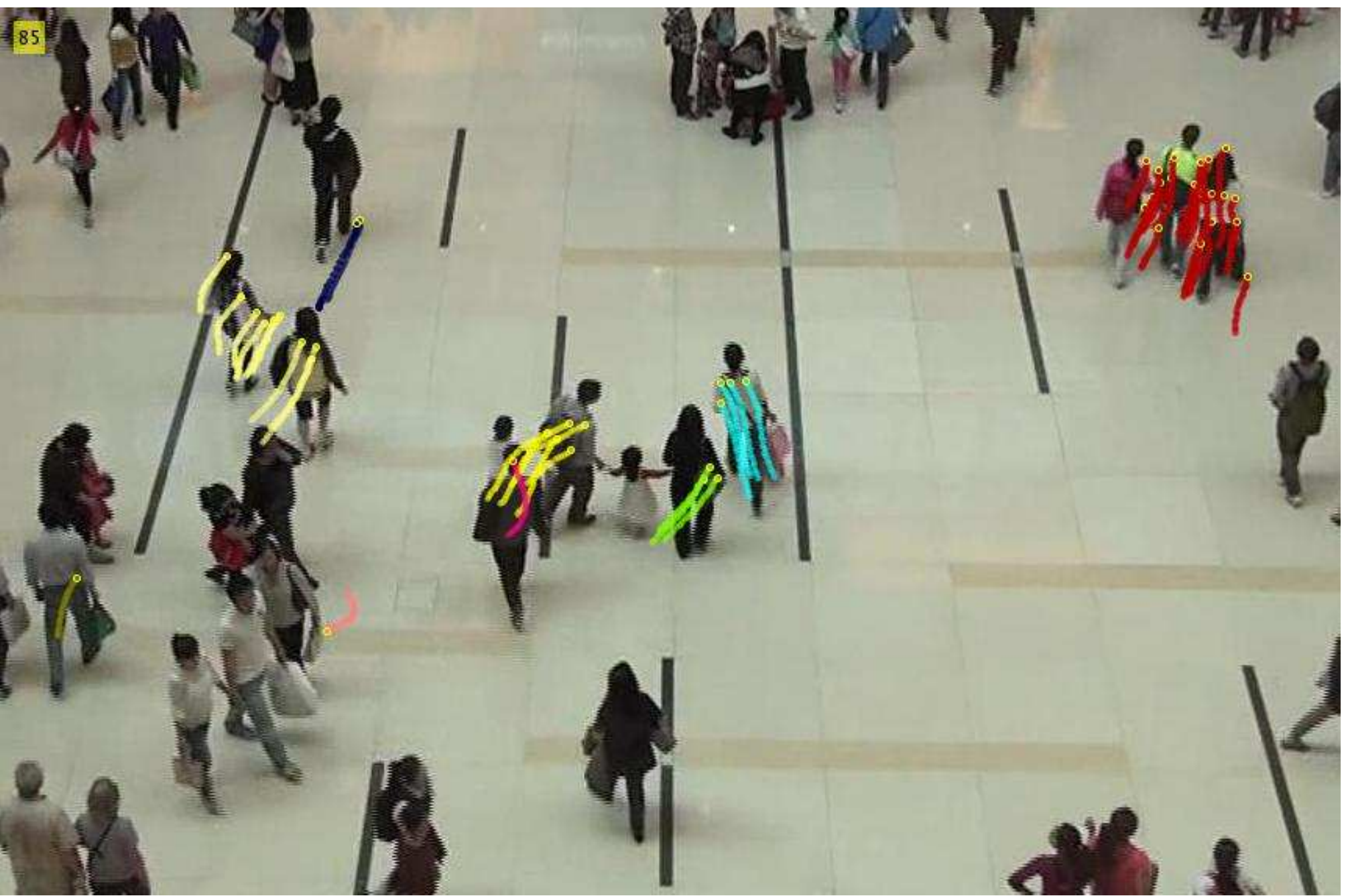}} \\
\vspace{-0.25cm}
	\clearsubcaptcounter
	\subfloat{\includegraphics[scale=0.18]{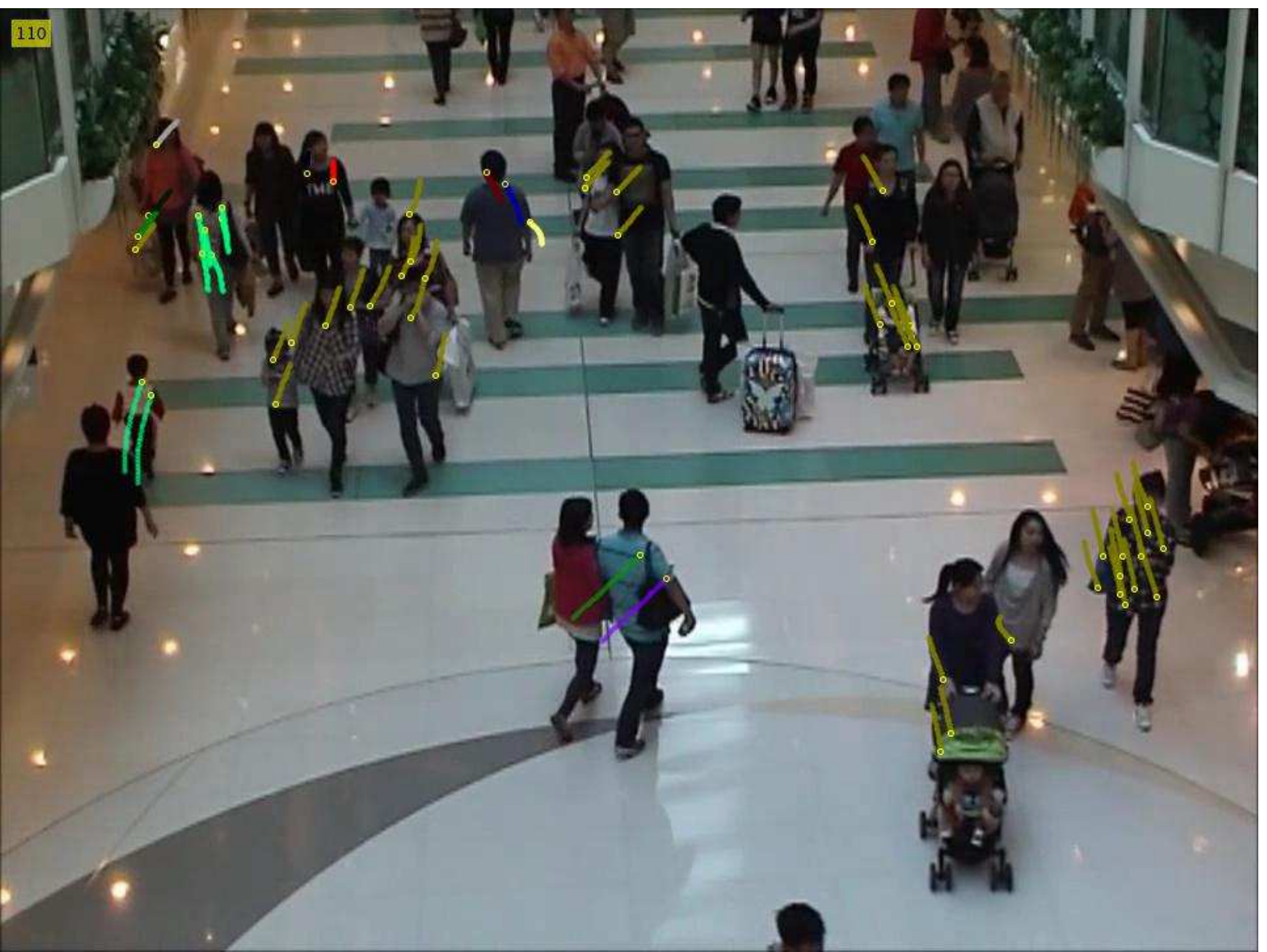}}\hspace{0.00001cm}
	\clearsubcaptcounter
	\subfloat{\includegraphics[scale=0.18]{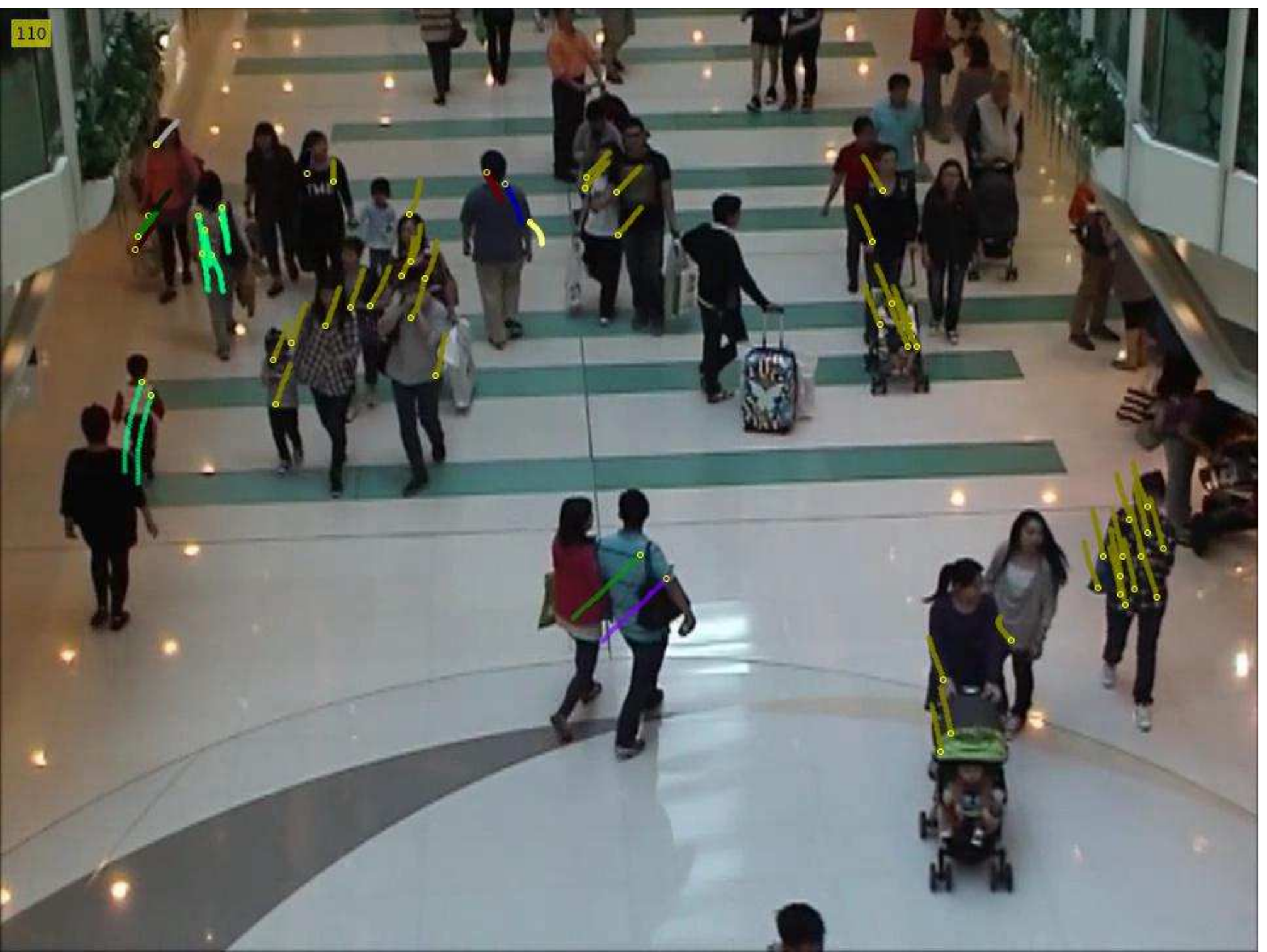}}\hspace{0.00001cm}
	\clearsubcaptcounter
	\subfloat{\includegraphics[scale=0.18]{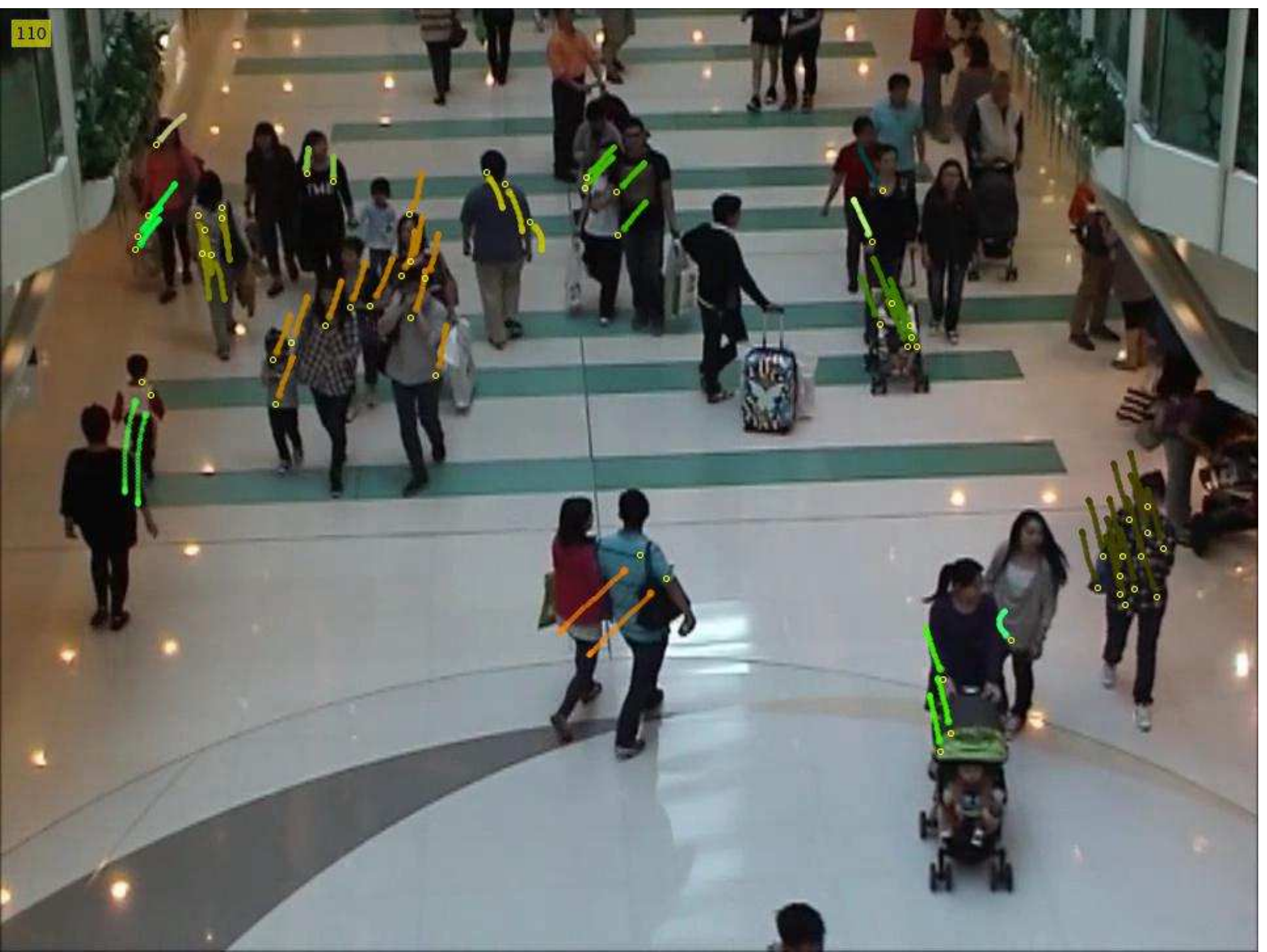}}\hspace{0.00001cm}
	\clearsubcaptcounter
	\subfloat{\includegraphics[scale=0.18]{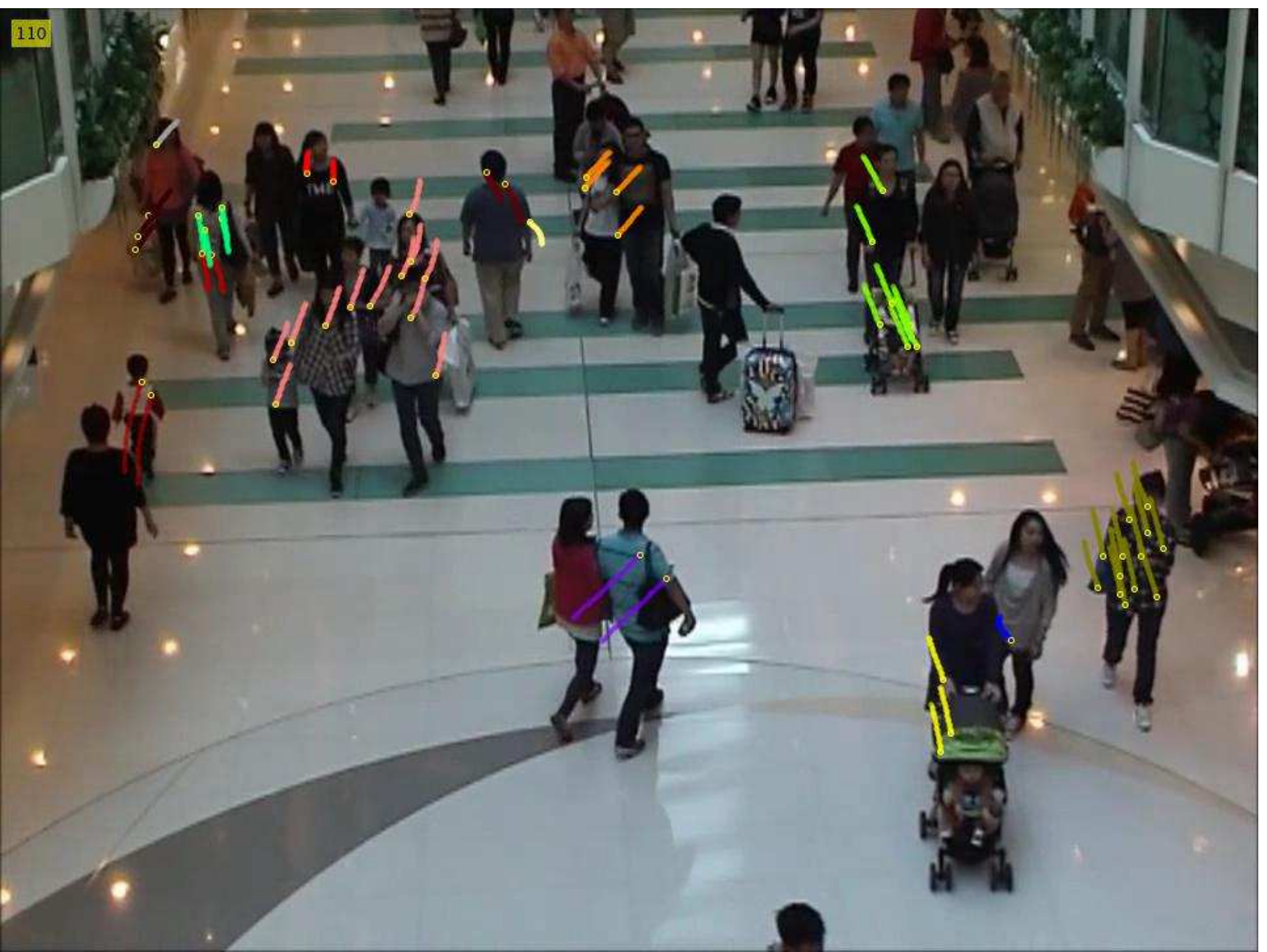}} \\
	\clearsubcaptcounter
\vspace{-0.25cm}
	\subfloat{\includegraphics[scale=0.18]{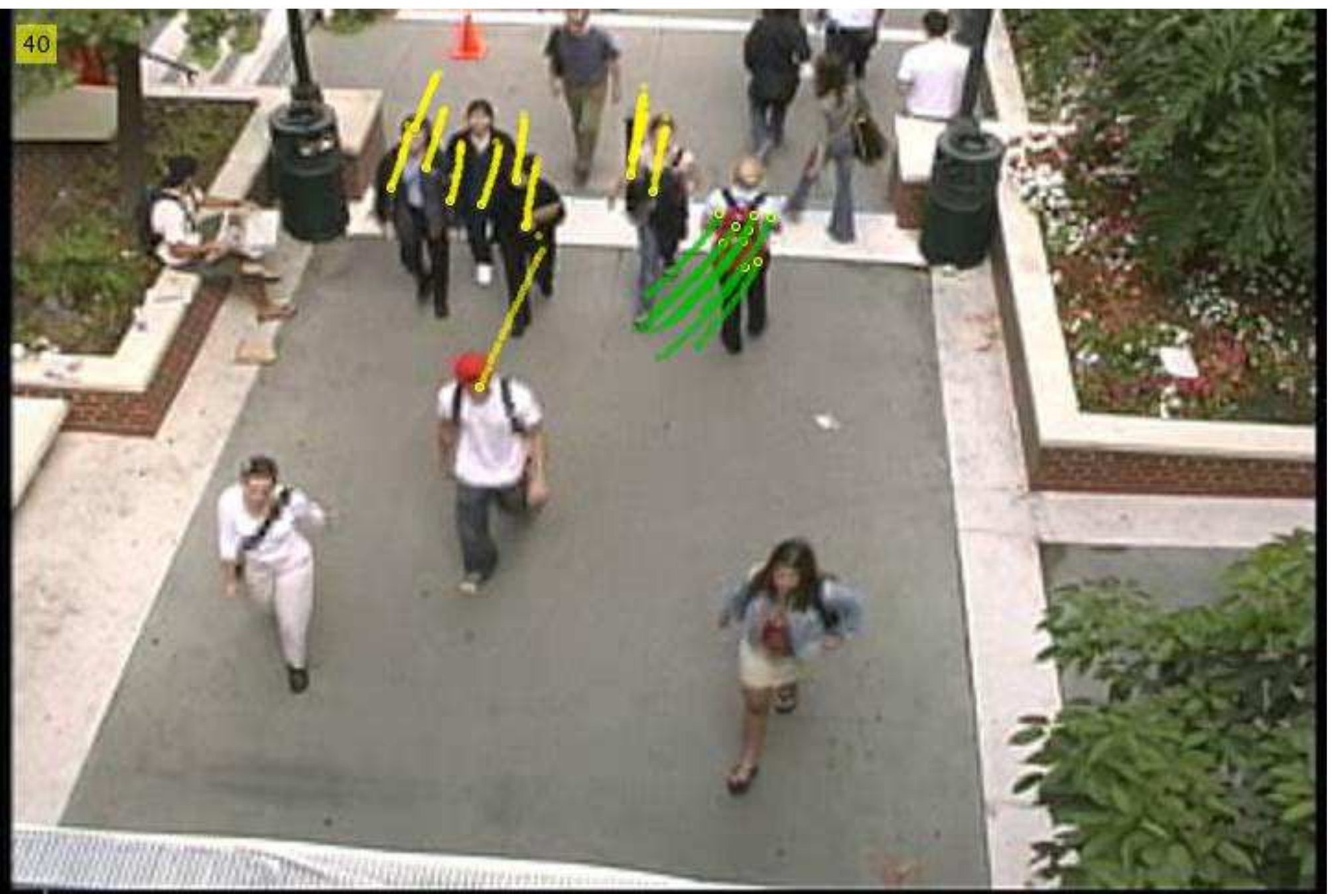}}\hspace{0.00001cm}
	\clearsubcaptcounter
	\subfloat{\includegraphics[scale=0.18]{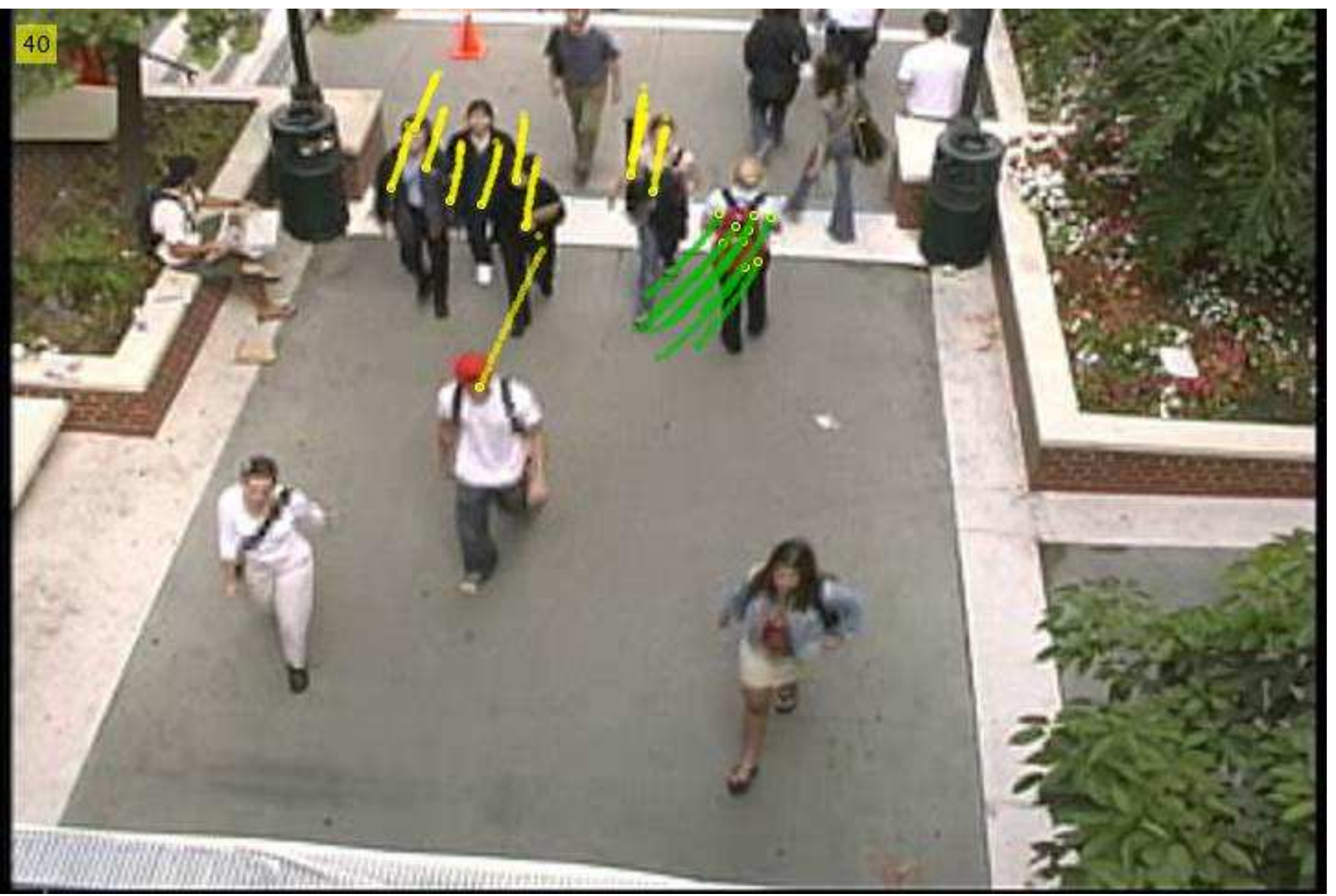}}\hspace{0.00001cm}
	\clearsubcaptcounter
	\subfloat{\includegraphics[scale=0.18]{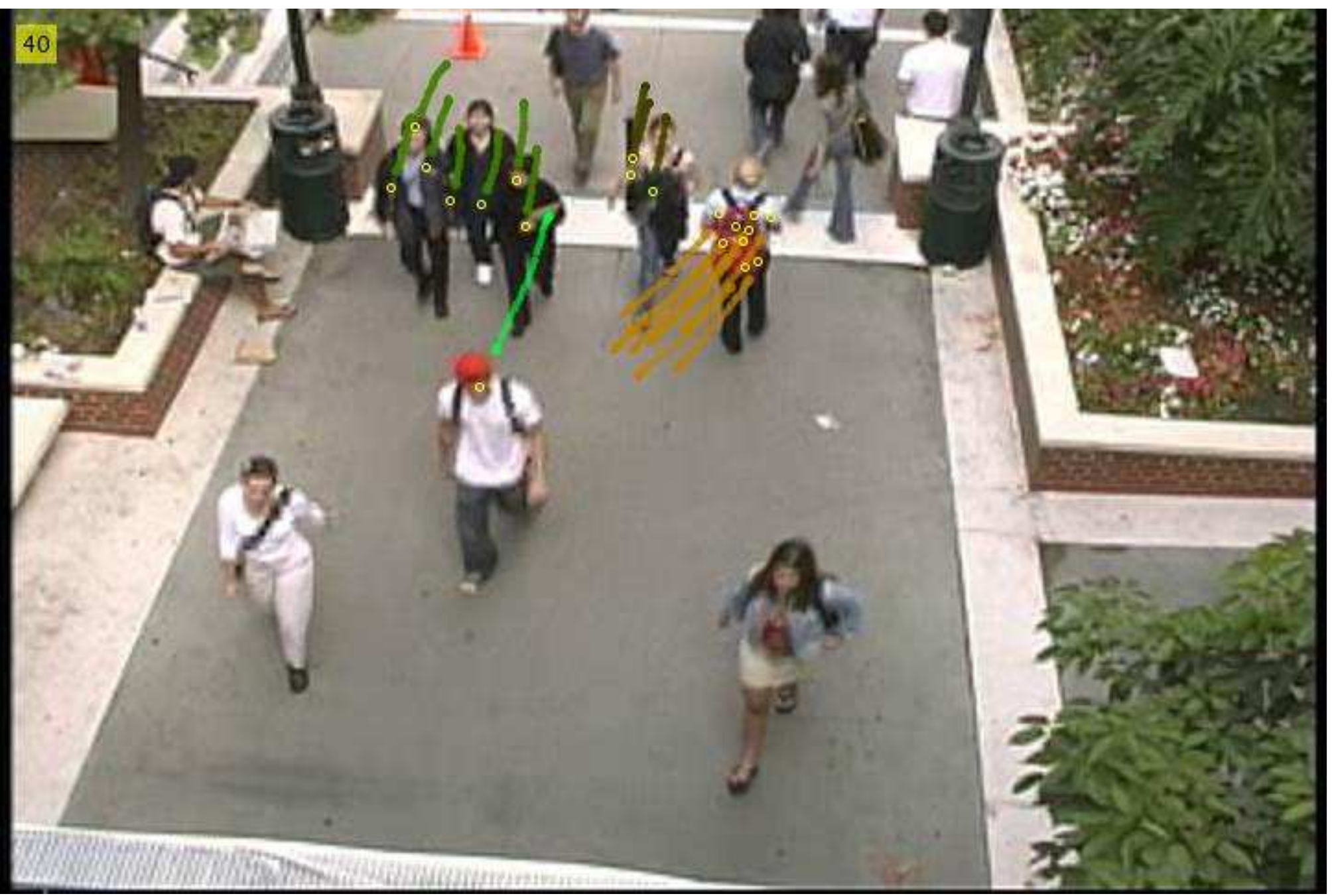}}\hspace{0.00001cm}
	\clearsubcaptcounter
	\subfloat{\includegraphics[scale=0.18]{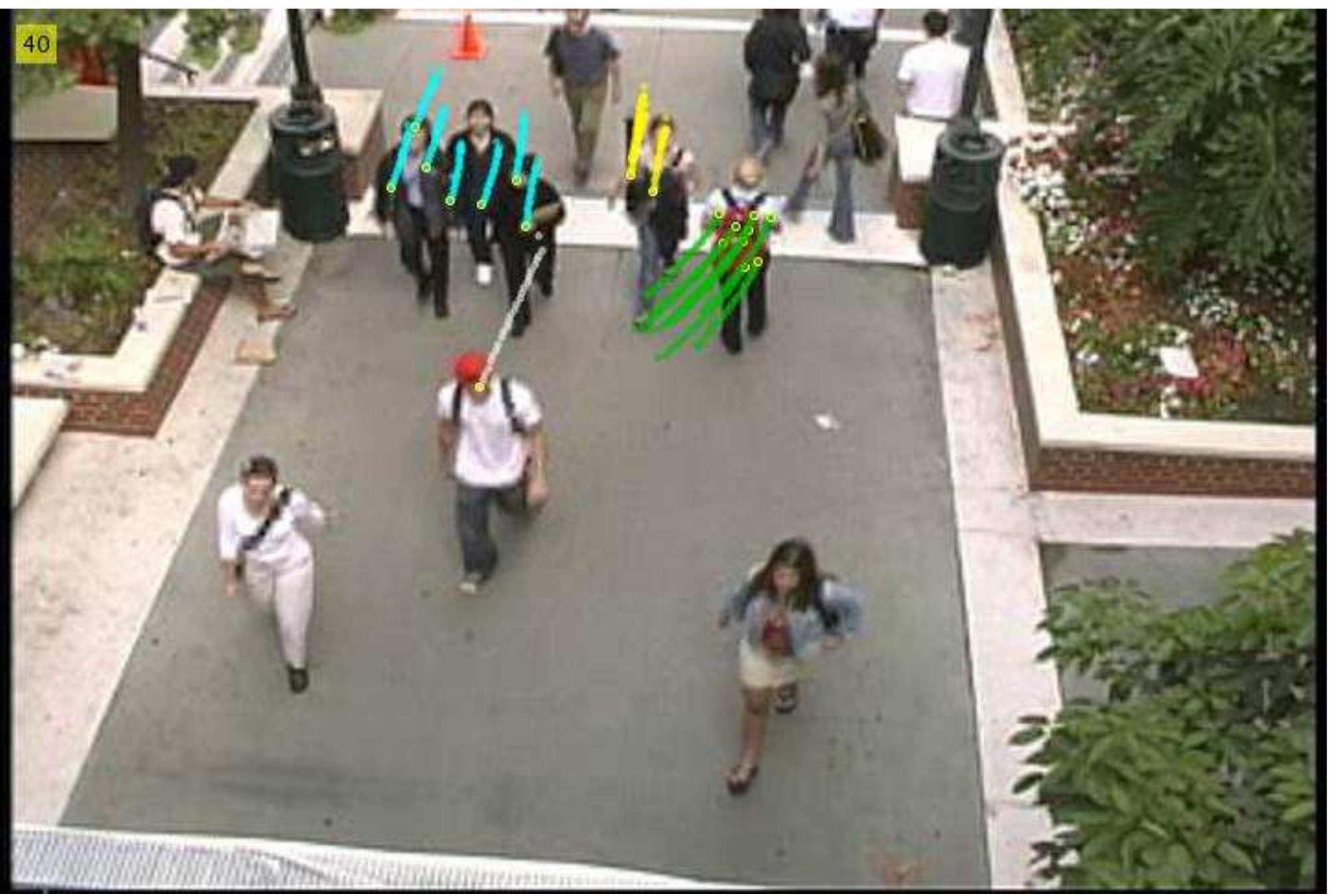}} \\
	\clearsubcaptcounter
\vspace{-0.25cm}
	\subfloat{\includegraphics[scale=0.18]{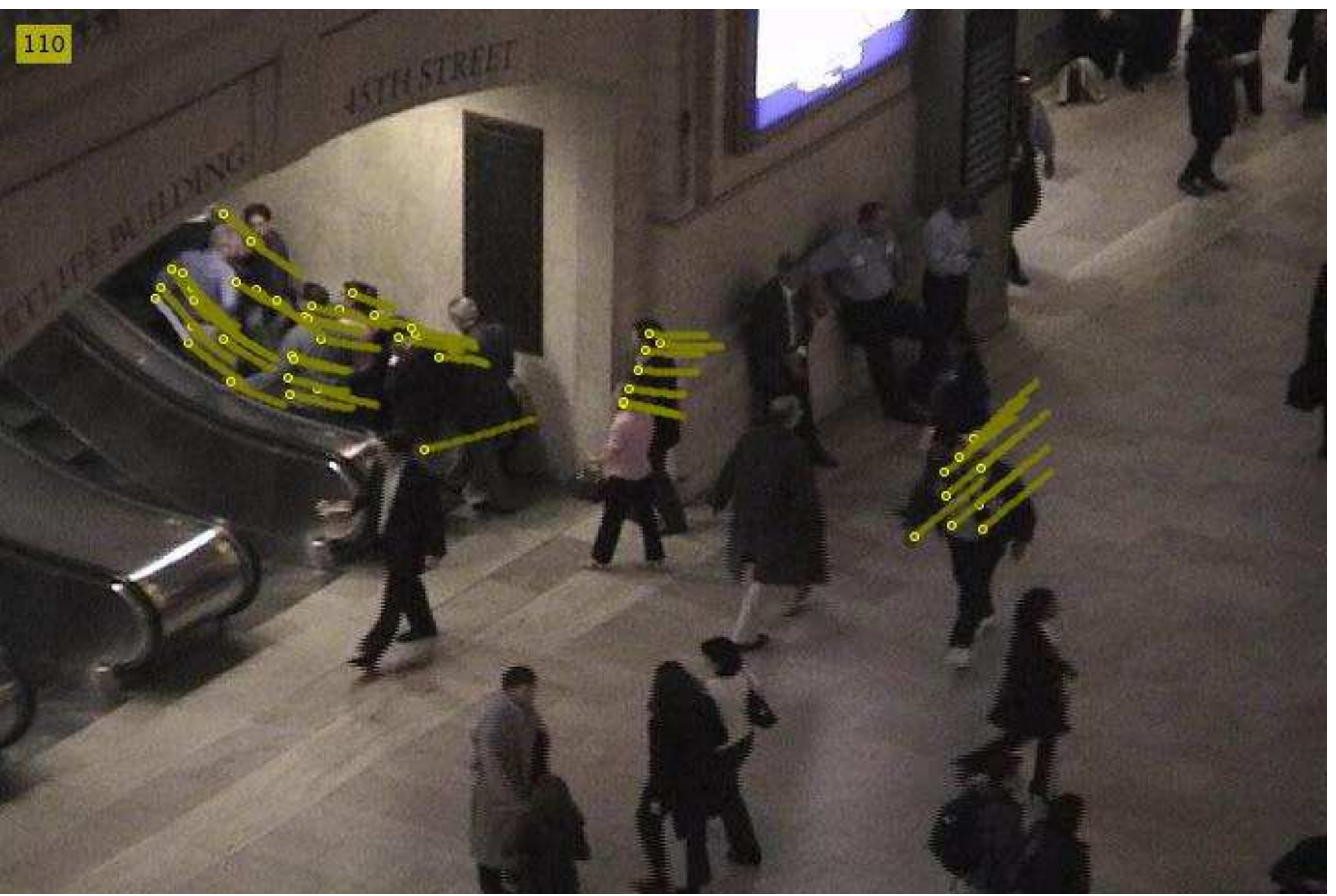}}\hspace{0.00001cm}
	\clearsubcaptcounter
	\subfloat{\includegraphics[scale=0.18]{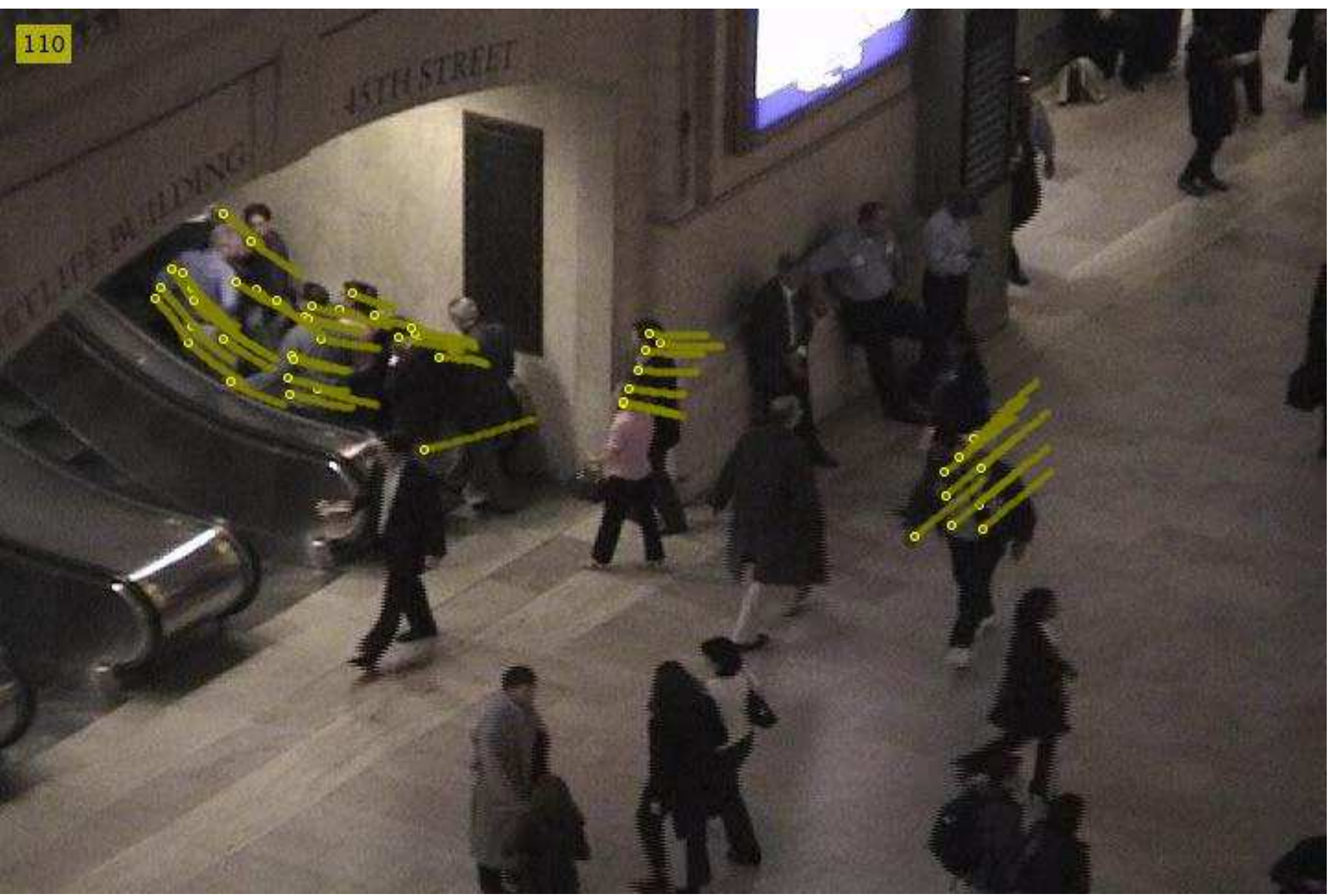}}\hspace{0.00001cm}
	\clearsubcaptcounter
	\subfloat{\includegraphics[scale=0.18]{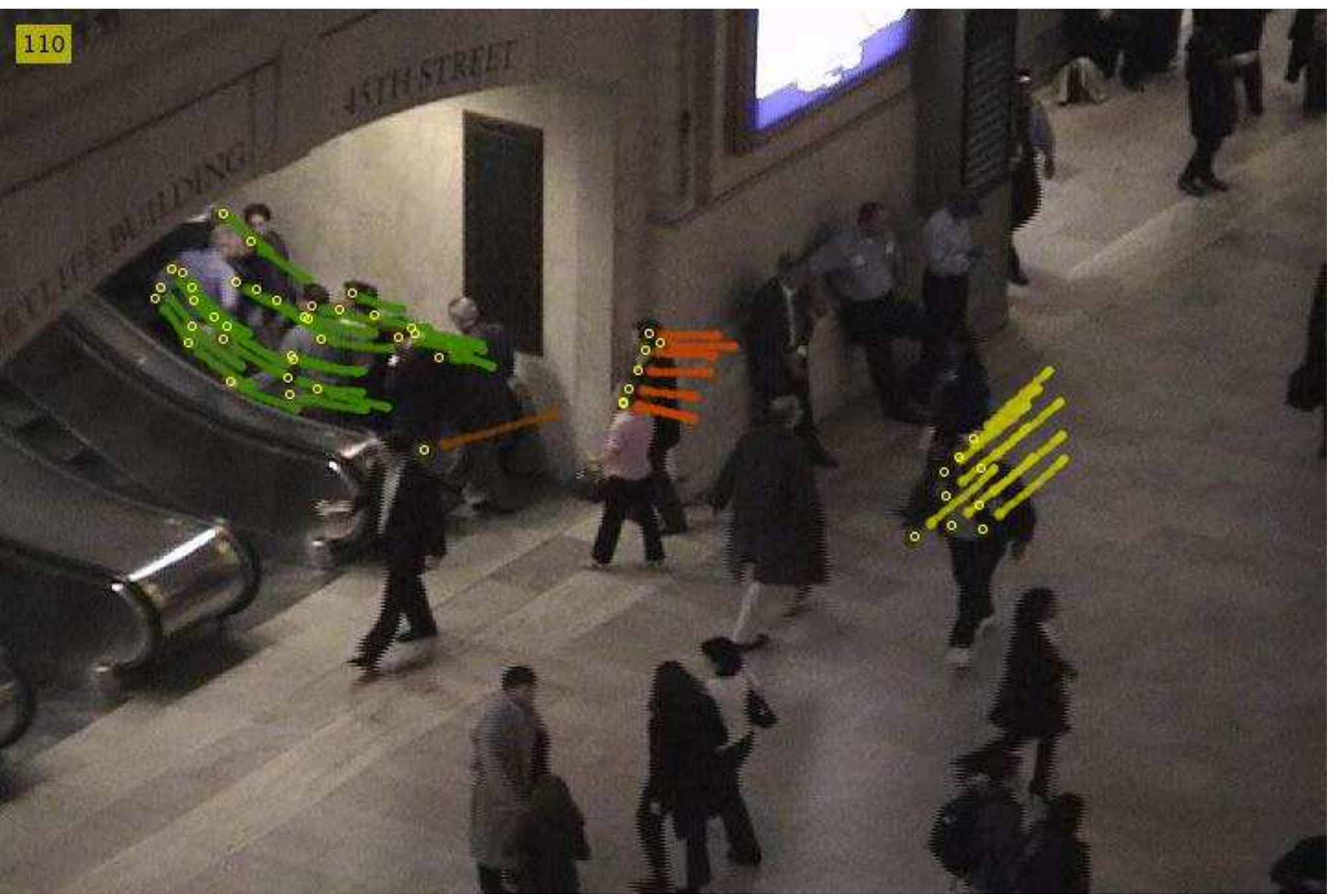}}\hspace{0.00001cm}
	\clearsubcaptcounter
	\subfloat{\includegraphics[scale=0.18]{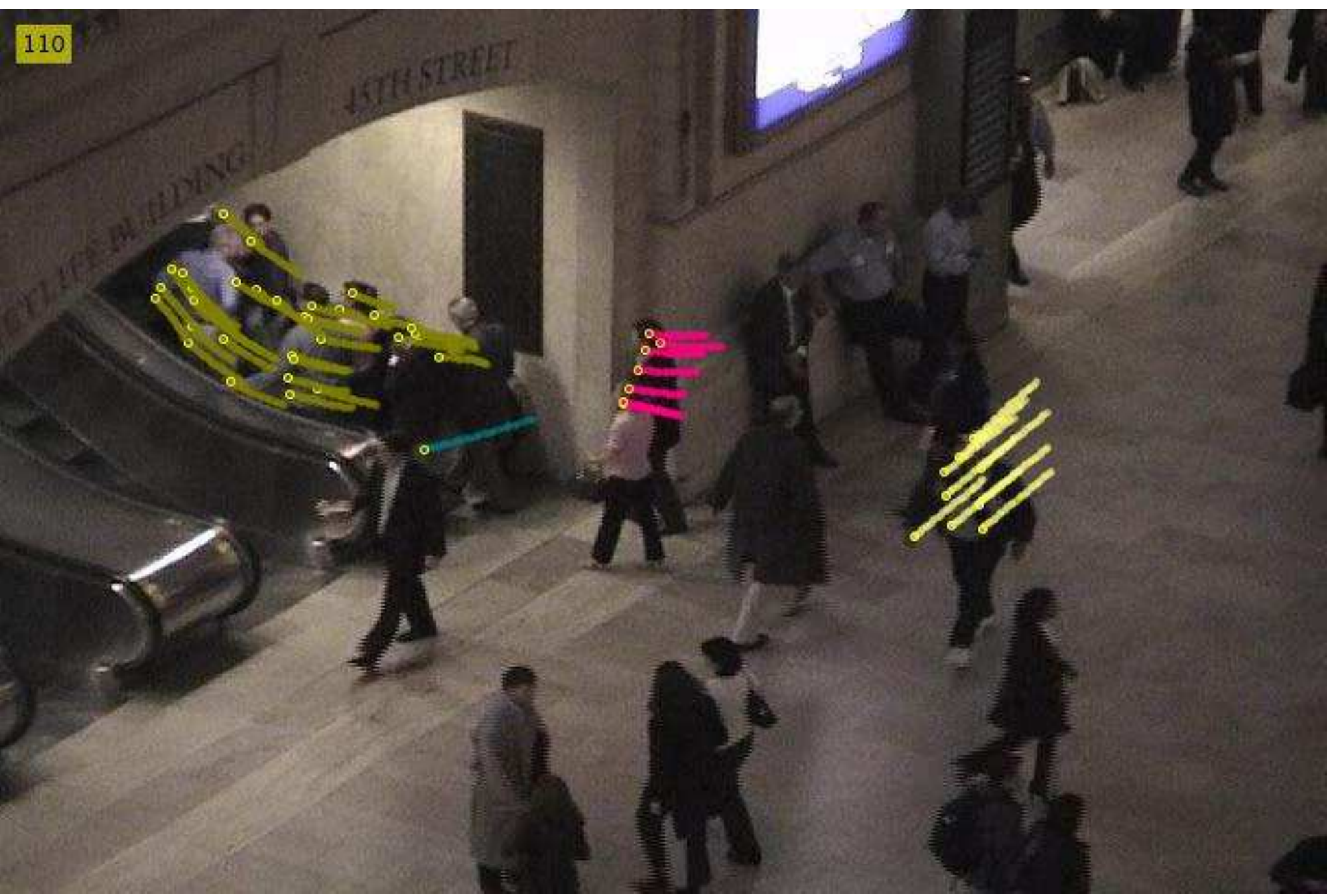}}\\
	\clearsubcaptcounter
\vspace{-0.25cm}
	\subfloat{\includegraphics[scale=0.18]{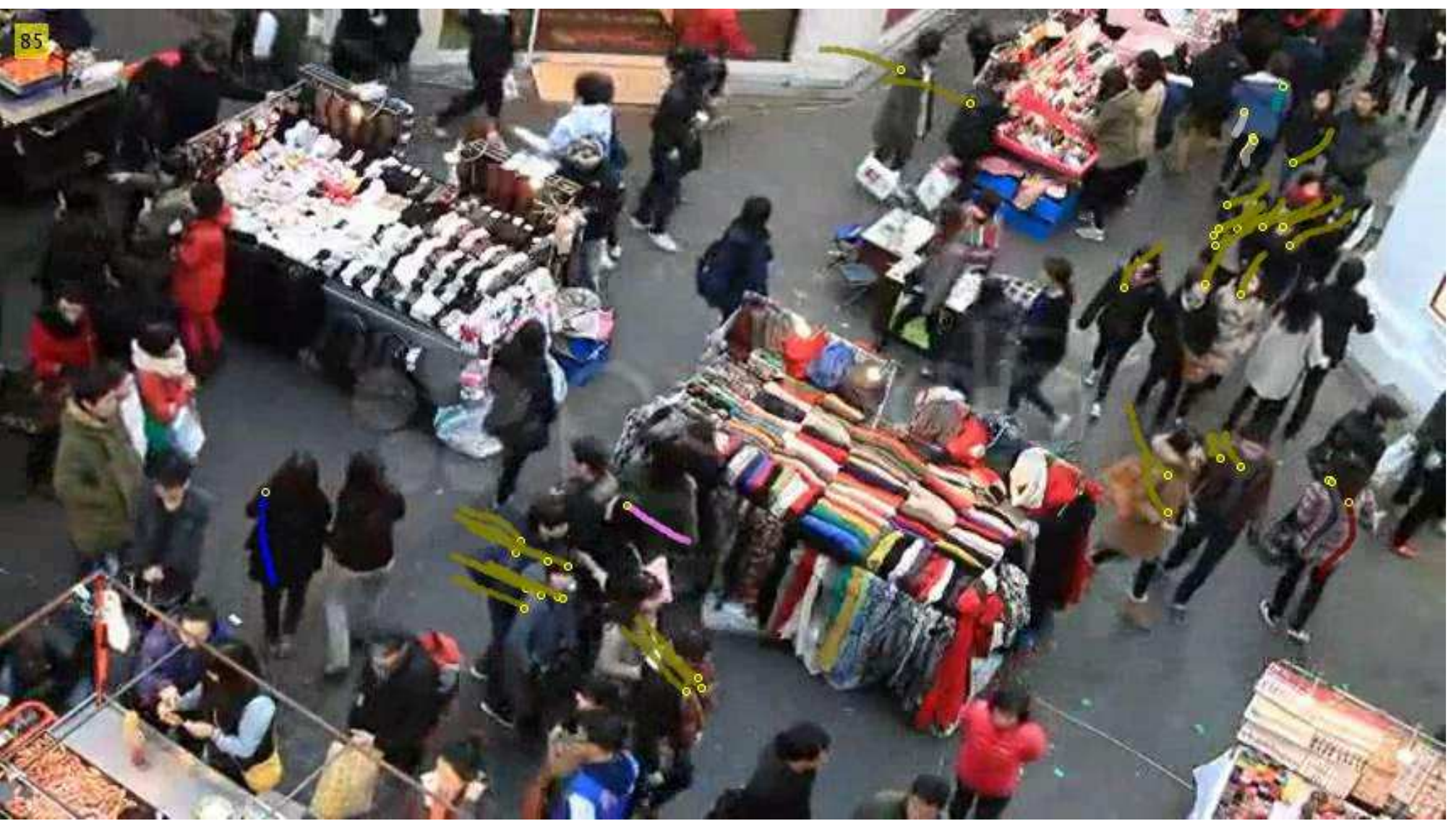}}\hspace{0.00001cm}
	\clearsubcaptcounter
	\subfloat{\includegraphics[scale=0.18]{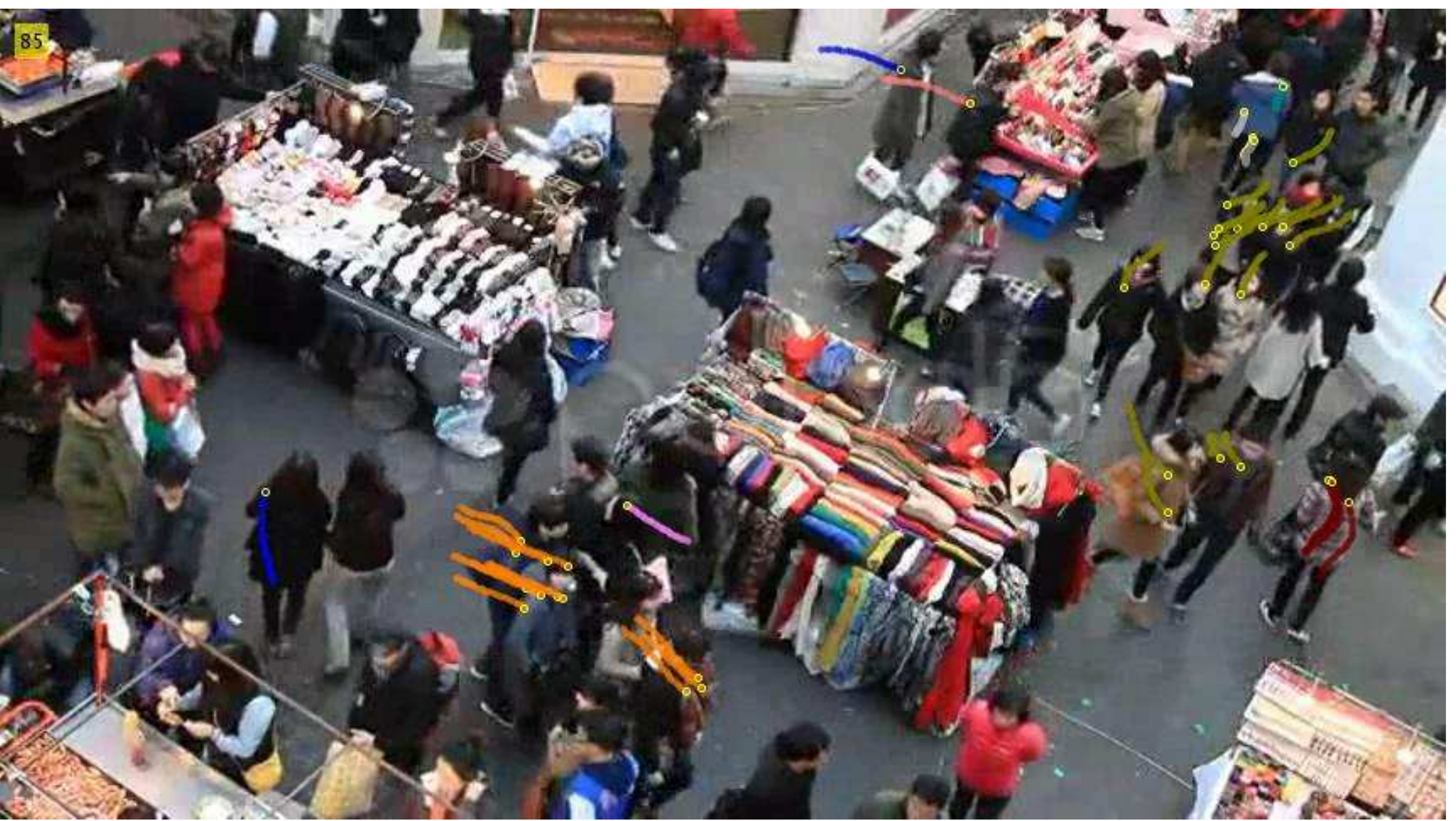}}\hspace{0.00001cm}
	\clearsubcaptcounter
	\subfloat{\includegraphics[scale=0.18]{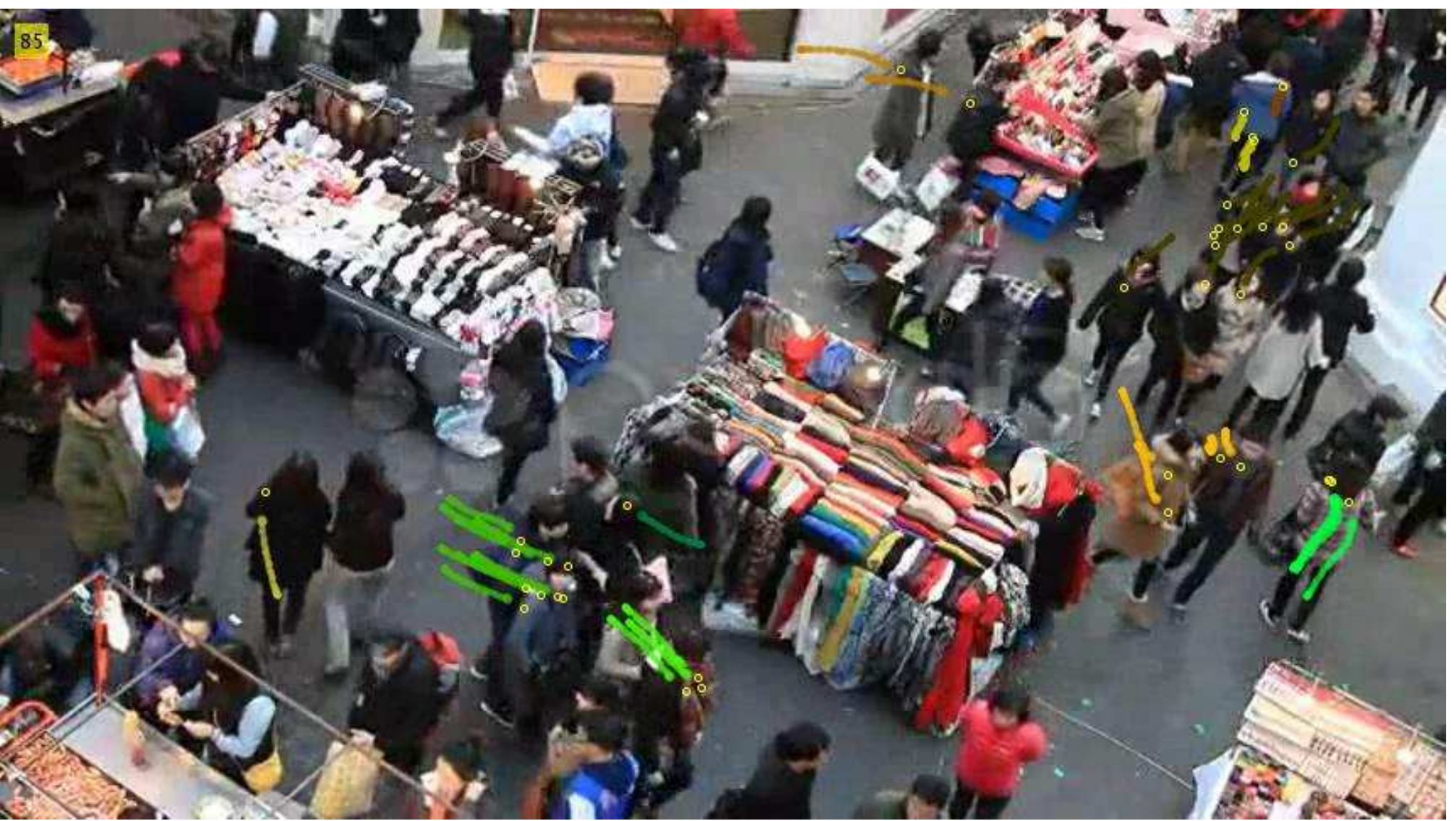}}\hspace{0.00001cm}
	\clearsubcaptcounter
	\subfloat{\includegraphics[scale=0.18]{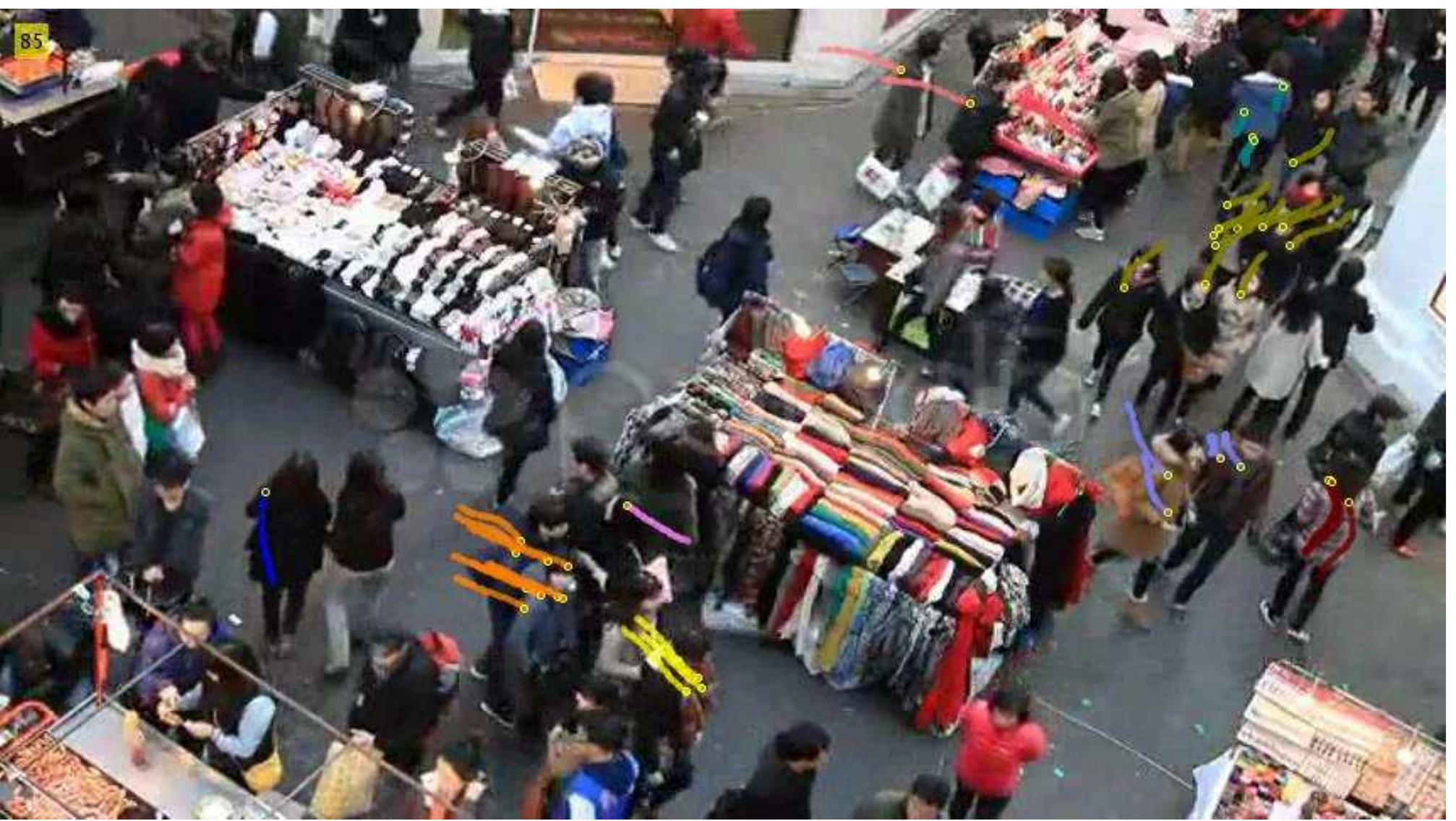}} \\
	\clearsubcaptcounter
\vspace{-0.25cm}
	\subfloat[CF]{\includegraphics[scale=0.18]{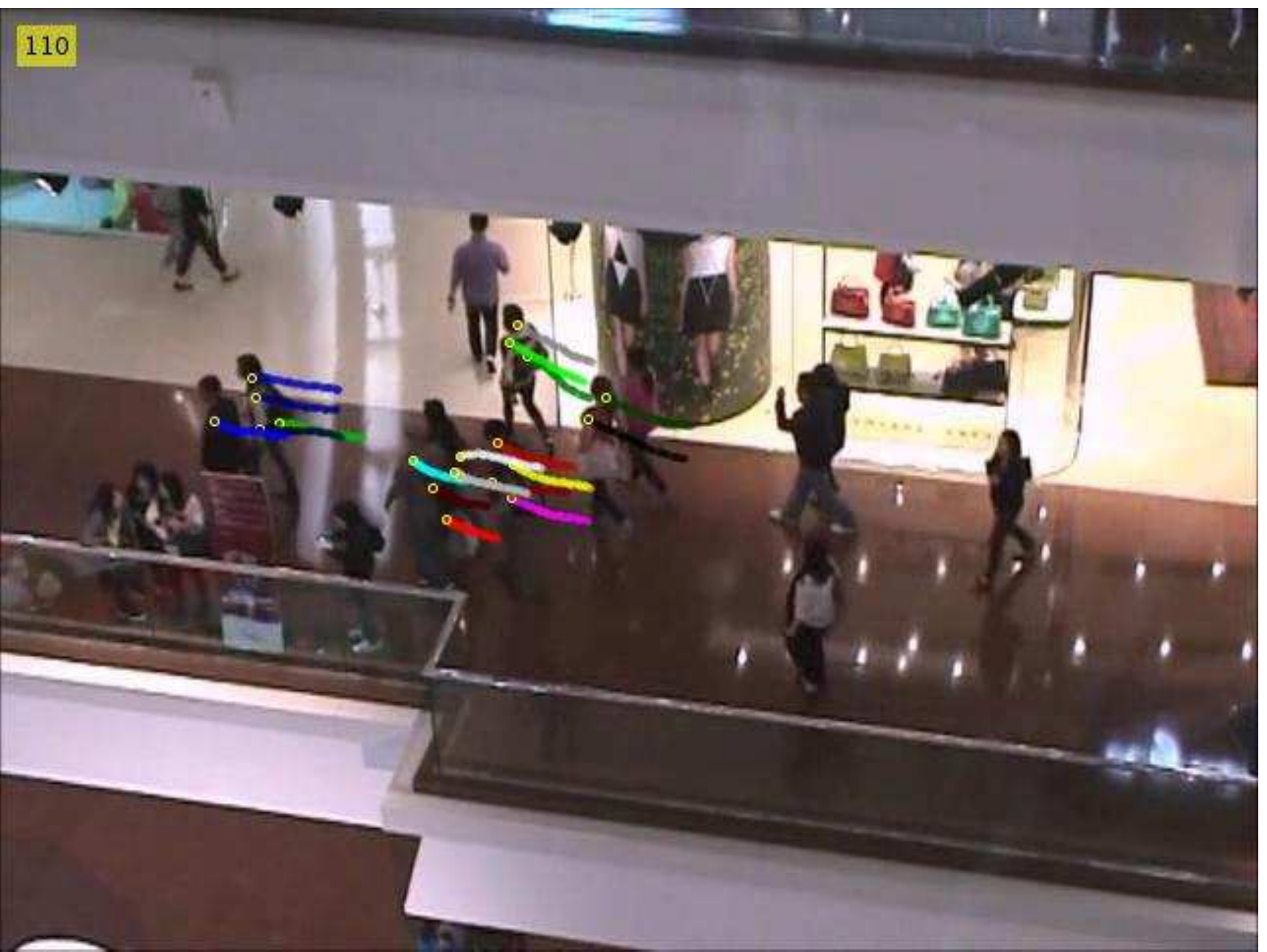}}\hspace{0.00001cm}
	\subfloat[CT]{\includegraphics[scale=0.18]{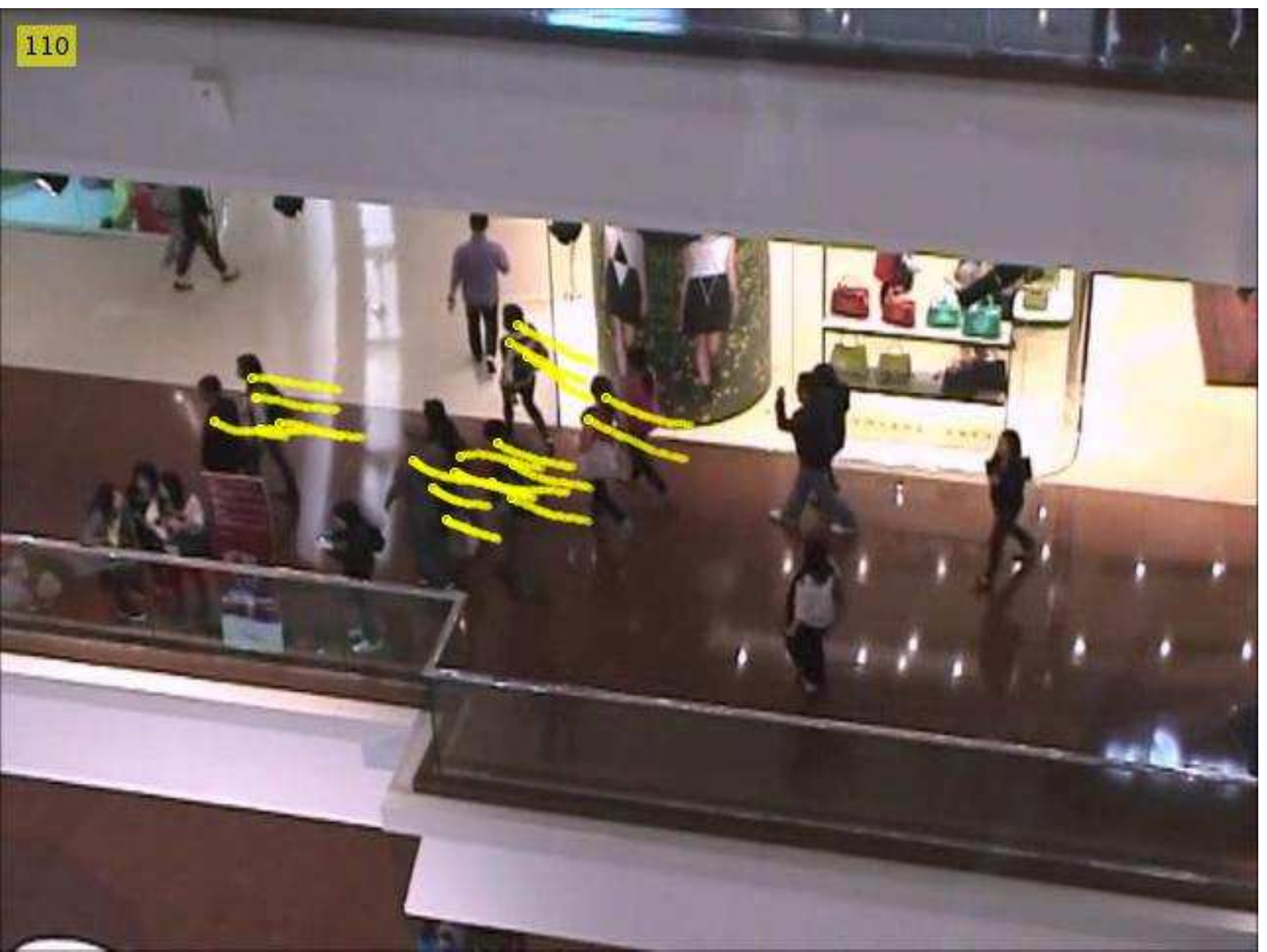}}\hspace{0.00001cm}  
	\subfloat[P]{\includegraphics[scale=0.18]{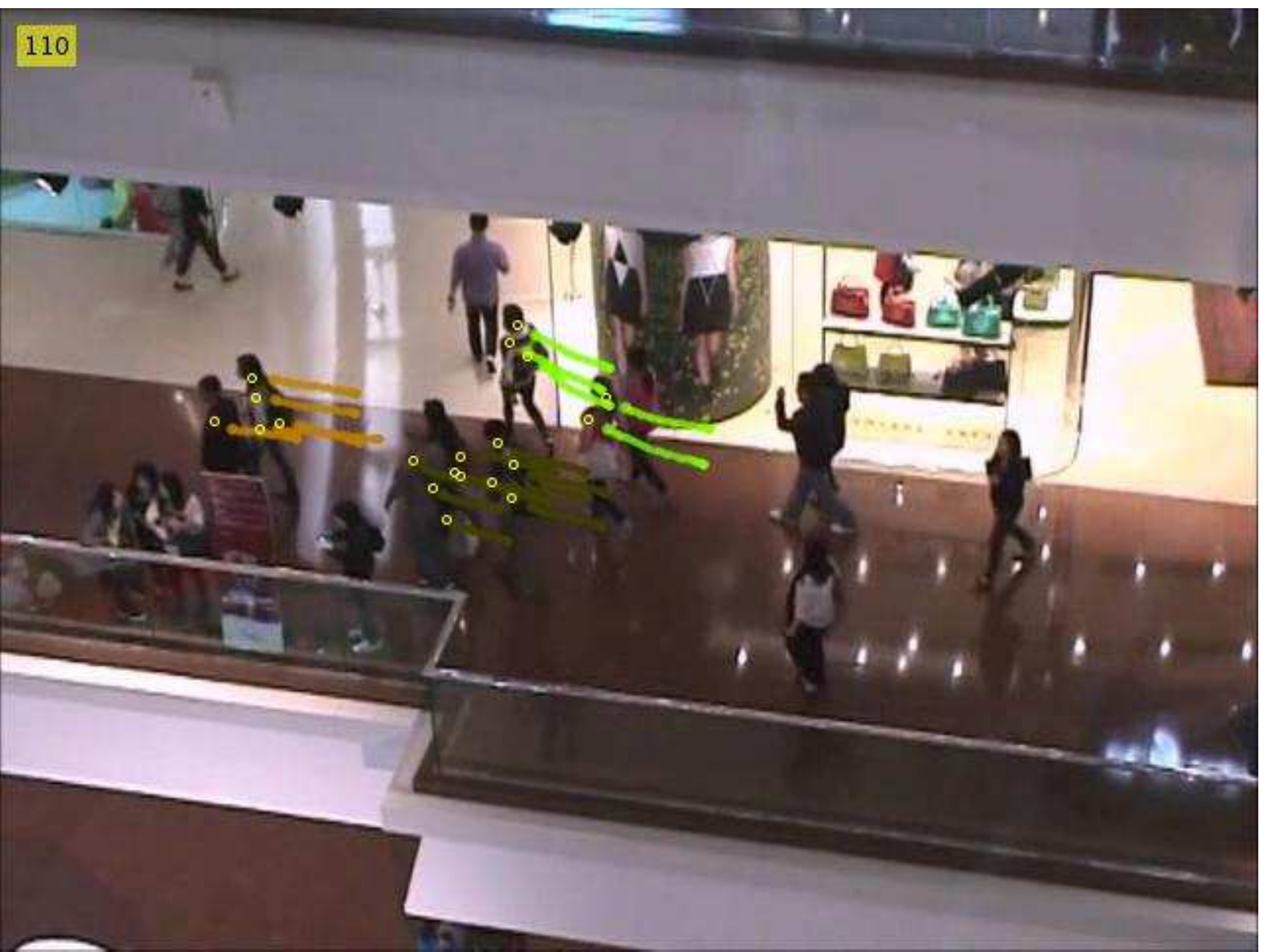}}\hspace{0.00001cm}
	\subfloat[GT]{\includegraphics[scale=0.18]{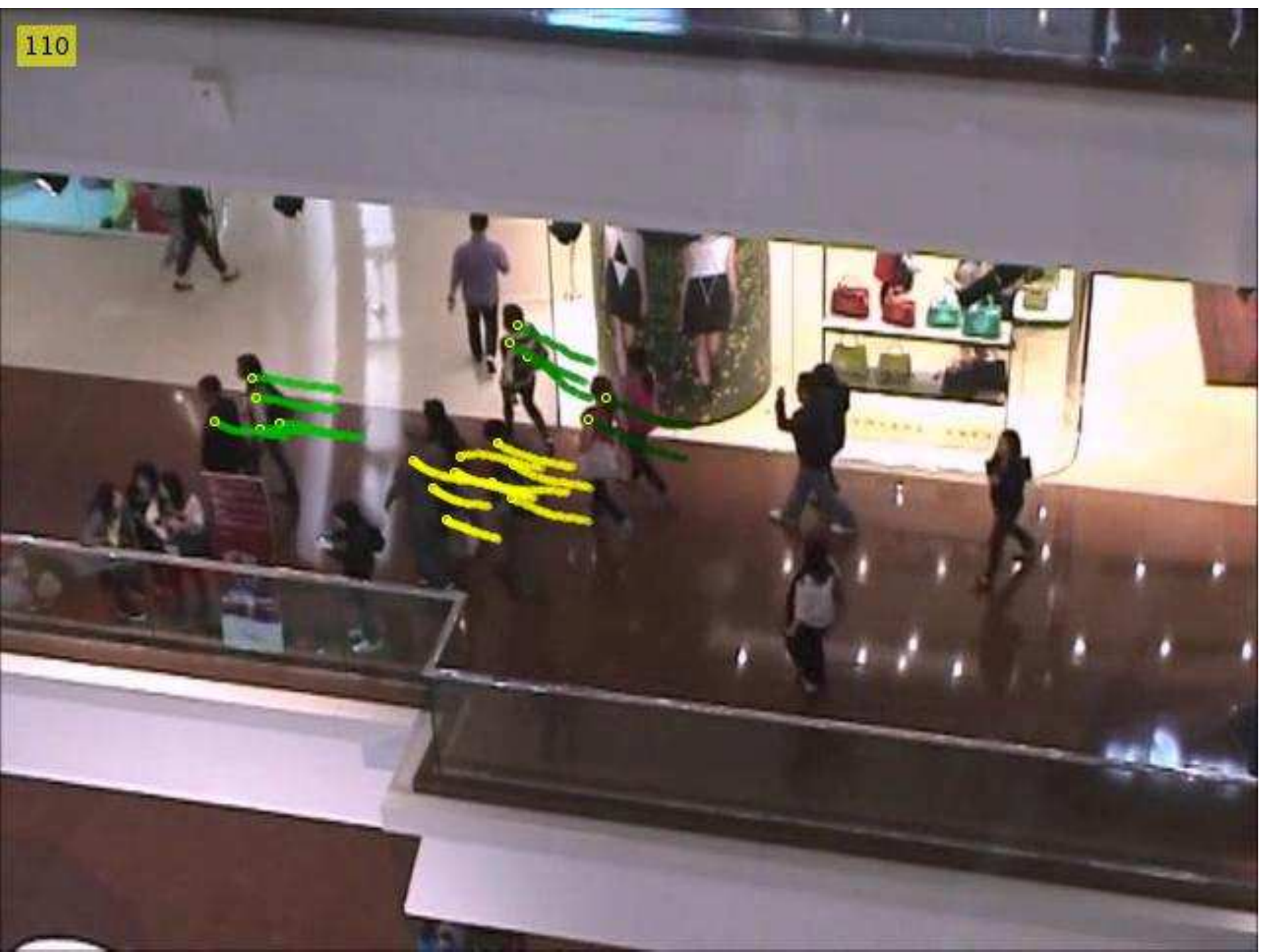}} \\
	\caption{Comparison of group detection results from Coherent Filtering \cite{cf} in column (a), Collective Transition \cite{scene} in column (b), our proposed method in column (c) with the ground truth in column (d) for different types of scenes. Each group is represented by a different color. Best viewed in color and when zoomed.}
	\label{fig:compare_fig}
\end{figure*}

Fig.~\ref{fig:compare_fig} demonstrates a visual comparison for different scenarios. Since Zhou \etal in~\cite{cf} find coherent motion patterns at one time and then update them over time, it is sensitive to tracking errors and has the possibility of accumulation of errors if any frame has tracking error. Shao \etal \cite{scene} assign every agent to a collective transition prior. They have spatial proximity constraint only at the initial time instant which might not be effective as time progresses. Their algorithm groups all the agents moving in the same direction giving less importance to their spatial relationships. This can be observed from the output figures in column (b) of Fig.~\ref{fig:compare_fig}. Further in $4^{th}$row, a person with red hat is moving faster than the group behind him but CT and CF fail to capture this difference in velocity while the proposed algorithm could capture it. The groups in last row have small changes in their directions of movement which is again not captured by these two methods while the proposed method detects such small changes.	 

We also compare the proposed group detection algorithm with the method of~\cite{scs} on the videos \textit{VEIIG}, \textit{student003} and \textit{eth}. To compare with ~\cite{scs}, we also use \textit{G}-MITRE precision \textbf{P} and recall \textbf{R} as proposed by them. Table~\ref{table:comp_scs} shows the quantitative results that indicate an improved performance by the proposed method.

The proposed algorithm outperforms these state-of-the-art methods because it is more robust to tracking errors since we extract groups from the eigenvectors rather than directly using the tracklets. It is quite evident from Fig.~\ref{fig:compare_fig} where the tracklets for various agents are marked with different colors to indicate the group they belong to, that the proposed algorithm is able to detect agents in a group much better than the other existing methods. Also the proposed algorithm yields \textit{NMI} $= 0.92$, \textit{Purity} $= 0.94$ and \textit{RI} $=0.93$ on video clips from BEHAVE dataset whereas the corresponding measures for \cite{scene} and \cite{cf} have very low values (\textit{e.g.} \textit Purity for \textit{CF} is only $0.35$). It shows that these methods do not perform well in videos of a sparse crowd whereas the proposed method can also handle a sparse crowd effectively. 

\begin{figure*}
\centering
	\subfloat{\includegraphics[scale=0.14]{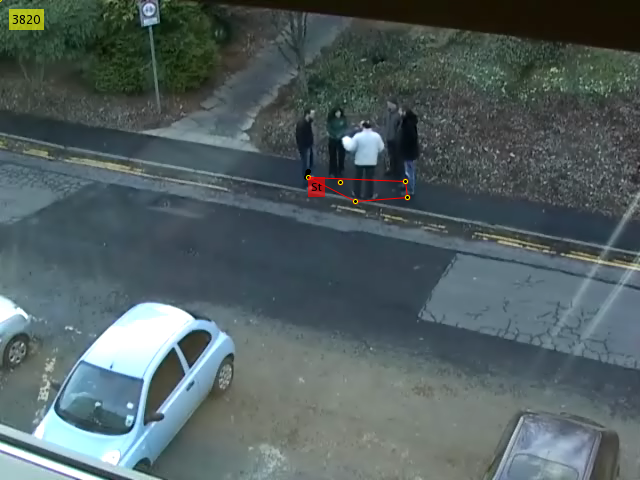}}\hspace{0.001cm}
\clearsubcaptcounter
	\subfloat{\includegraphics[scale=0.155]{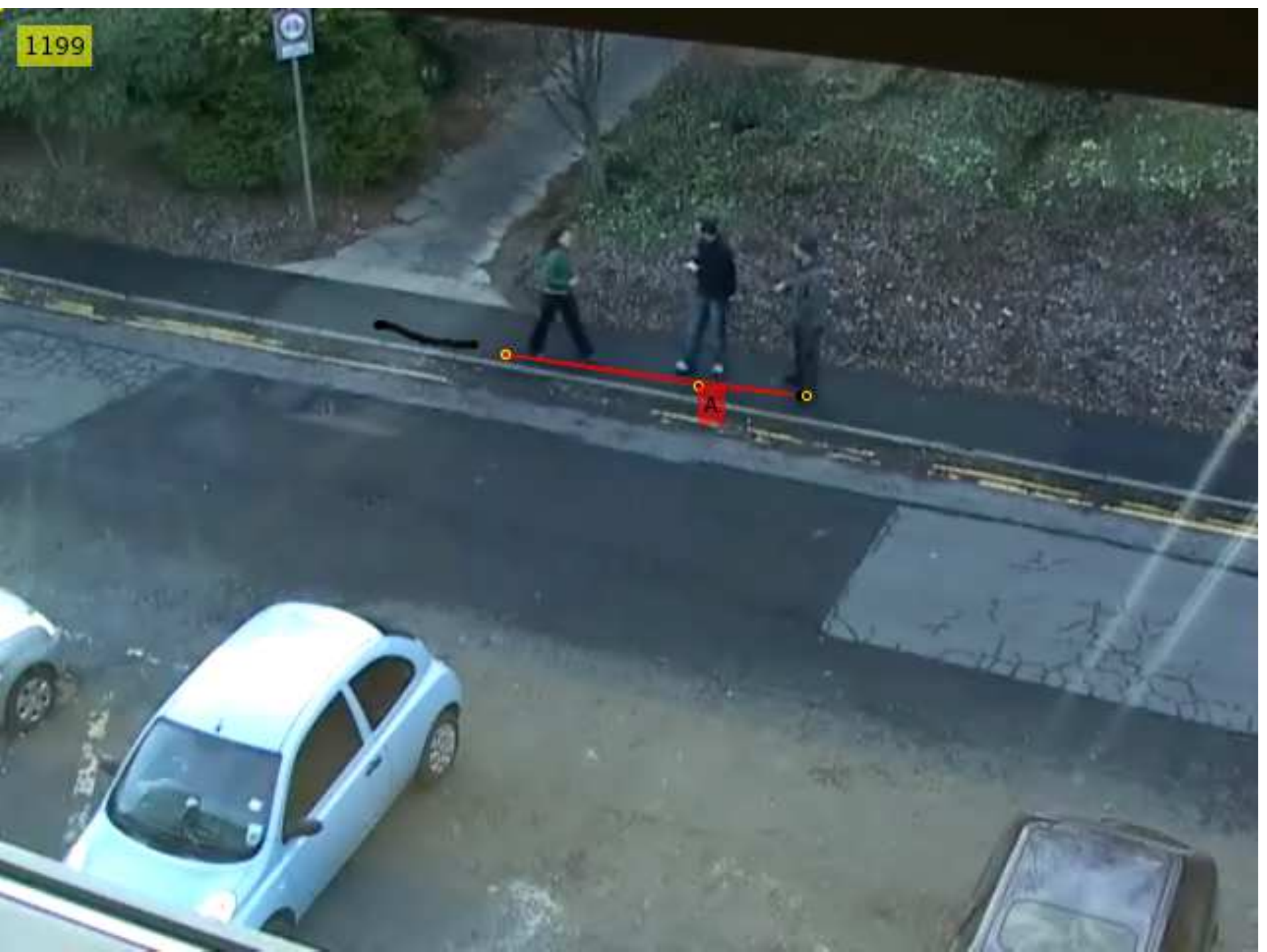}}\hspace{0.001cm}
\clearsubcaptcounter
	\subfloat{\includegraphics[scale=0.14]{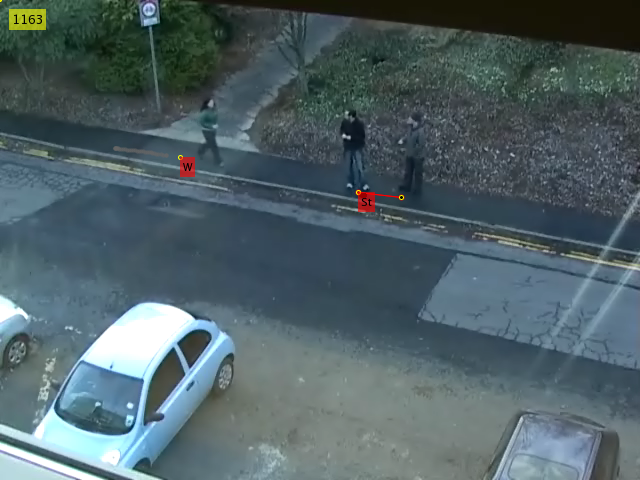}}\hspace{0.001cm}
\clearsubcaptcounter
	\subfloat{\includegraphics[scale=0.14]{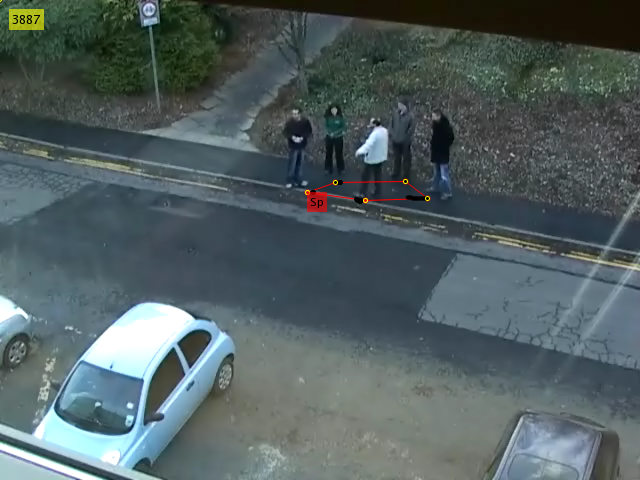}}\\
\clearsubcaptcounter
	\subfloat[stationary]{\includegraphics[scale=0.14]{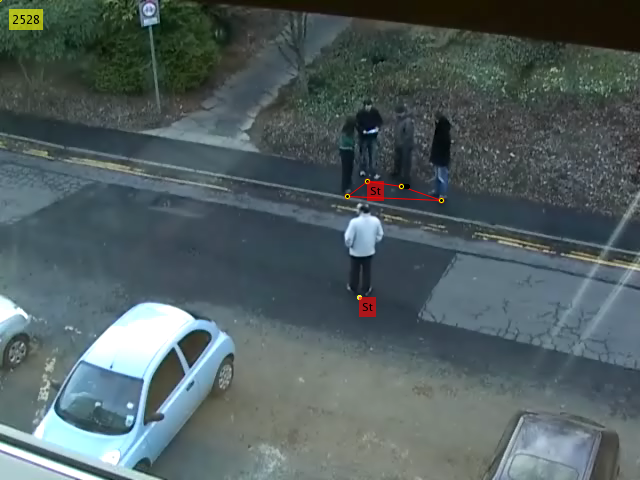}}\hspace{0.001cm}
	\subfloat[approaching]{\includegraphics[scale=0.155]{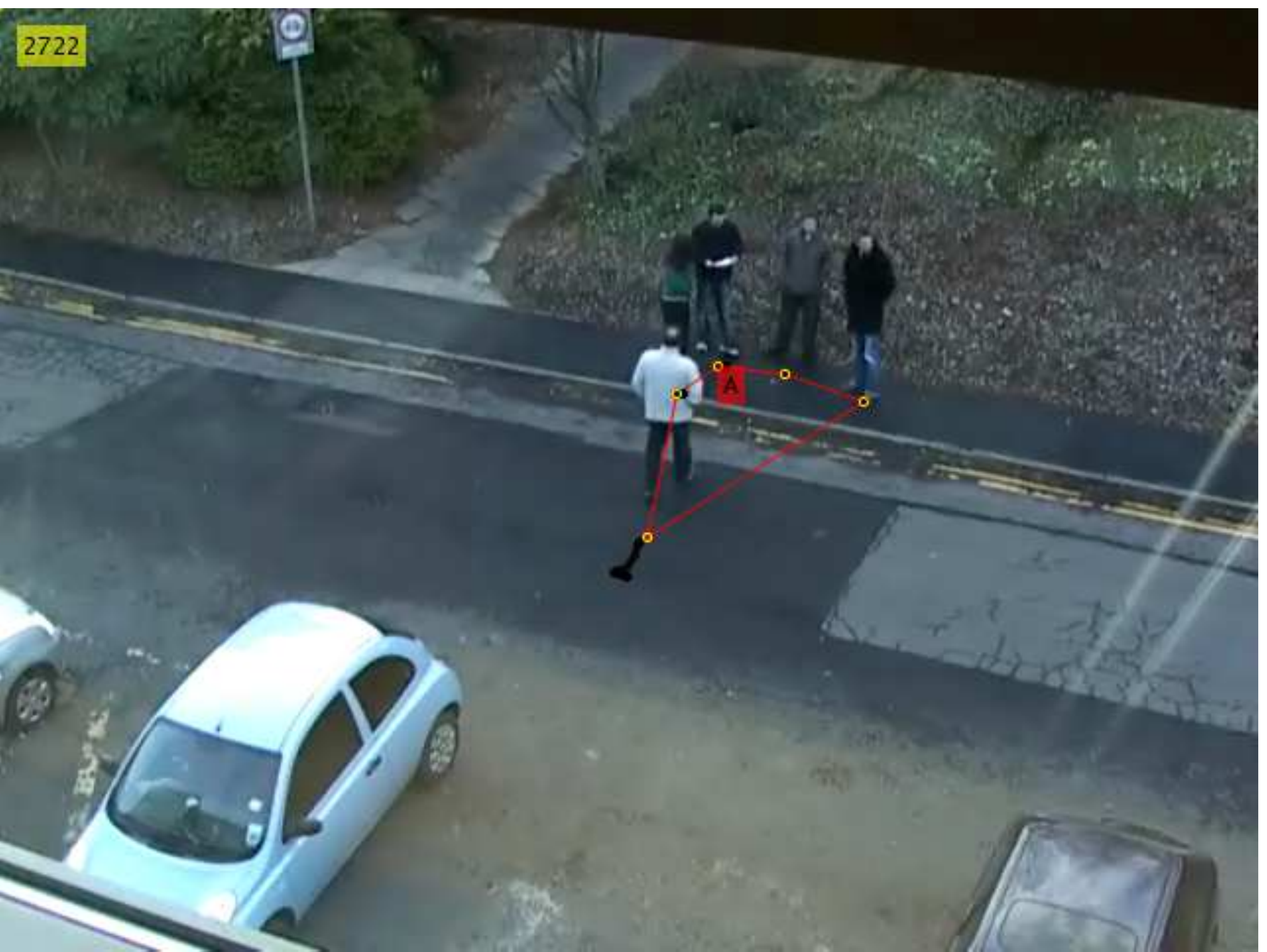}}\hspace{0.001cm}
	\subfloat[walking, stationary]{\includegraphics[scale=0.14]{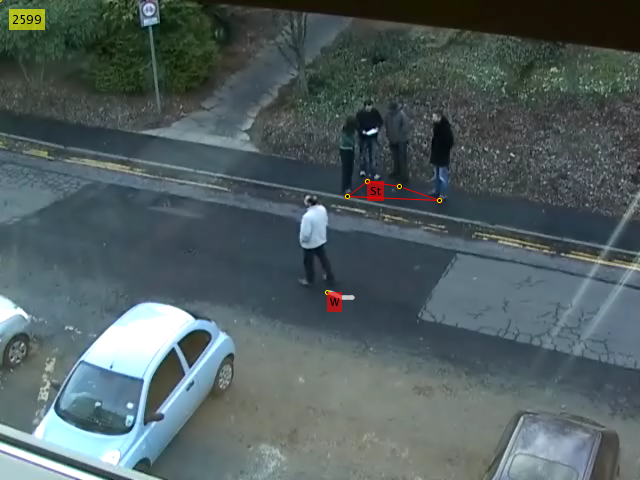}}\hspace{0.001cm}
	\subfloat[splitting]{\includegraphics[scale=0.14]{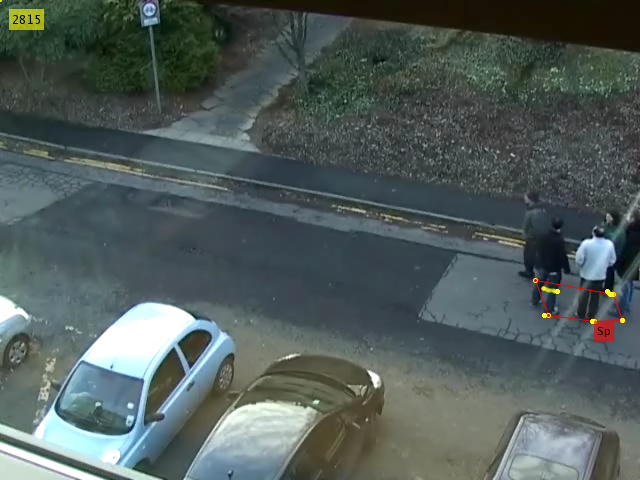}}\\	
\caption{Group activity results on BEHAVE dataset. Notation - \textbf{St}: Stationary, \textbf{A}: Approaching, \textbf{W}: Walking and \textbf{Sp}: Splitting. Best viewed in color and when zoomed.}
\label{fig:act_beh}
\end{figure*}

\begin{figure*}
\centering
\subfloat{\includegraphics[scale=0.23]{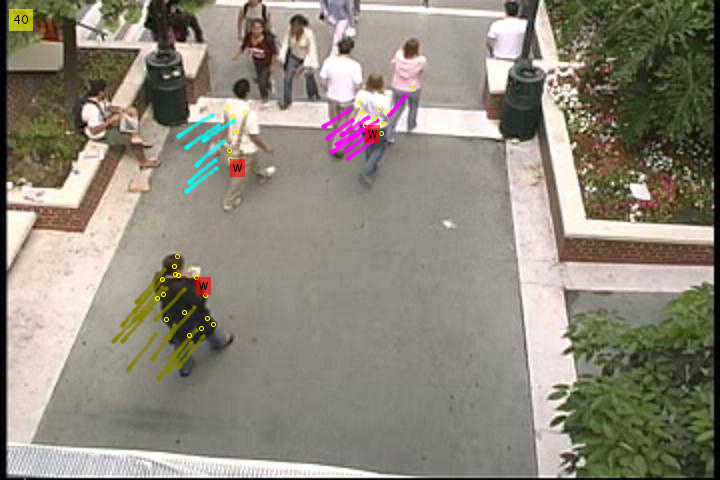}}\hspace{0.001cm}
\subfloat{\includegraphics[scale=0.195]{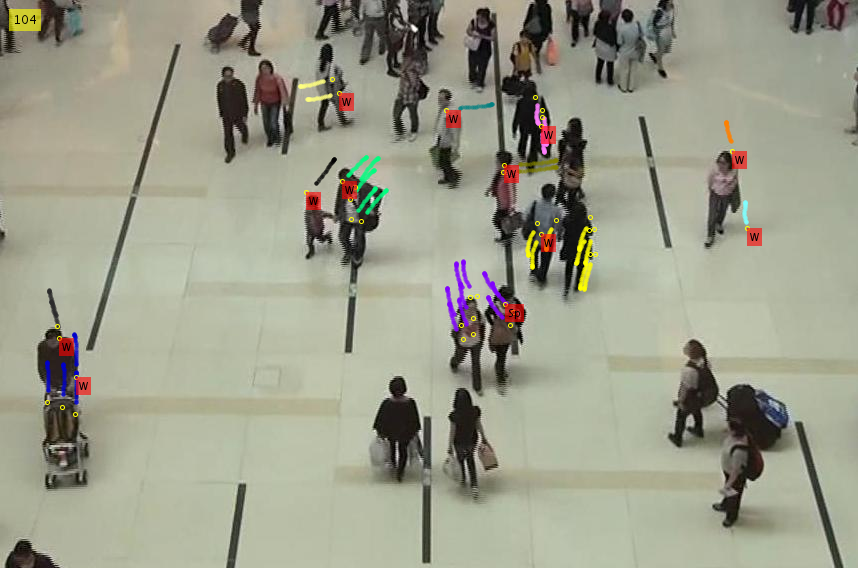}}\\
\subfloat{\includegraphics[scale=0.22]{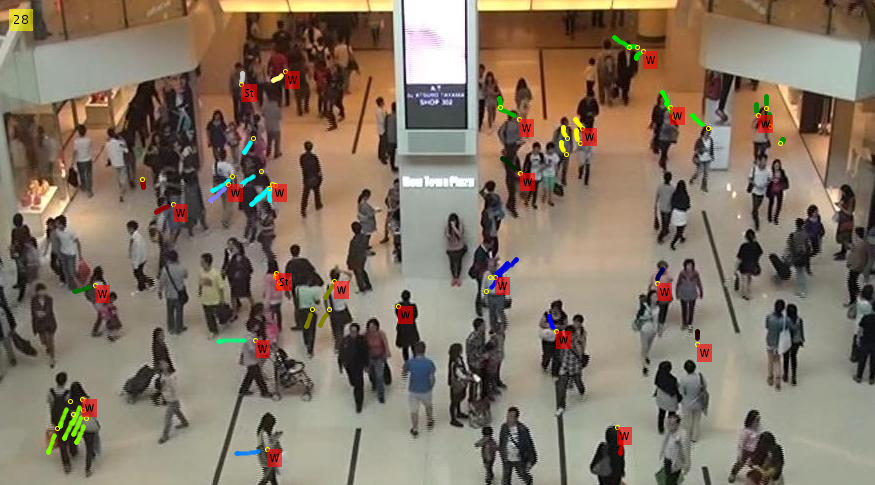}}\hspace{0.001cm}
\subfloat{\includegraphics[scale=0.295]{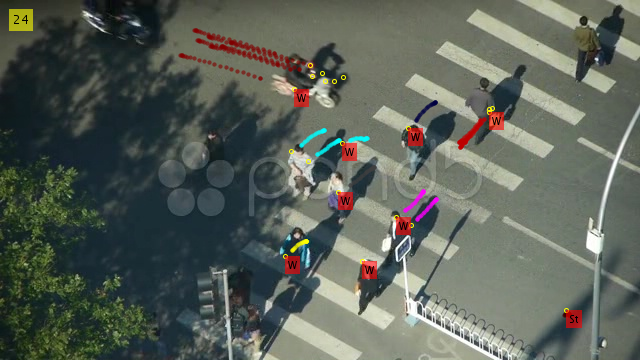}}\hspace{0.001cm}	
\caption{Group activity results on CUHK dataset. Same notation as in Fig.~\ref{fig:act_beh}. Best viewed in color and when zoomed.}
\label{fig:act_cukh}
\end{figure*}

\subsection{Group Activity Recognition}

\begin{figure}
\centering
\subfloat{\includegraphics[scale=0.4]{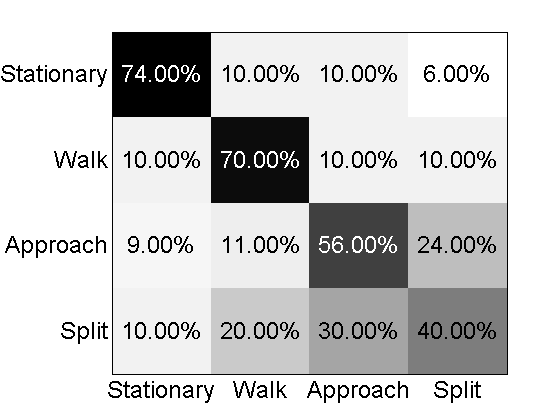}}
\caption{Confusion matrix for group activity on \textit{BEHAVE} dataset. The true classes are along the column and the predicted classes are along the row.}
\label{fig:grp_act}
\end{figure}

We use \textit{BEHAVE} and \textit{CUHK} datasets for testing the algorithm for group activity identification. Here, we have excluded the clips containing other activities such as fight. We compared the activity results with the ground truth at regular intervals. Table \ref{fig:grp_act} shows the confusion matrix for the proposed algorithm on \textit{BEHAVE} dataset. The algorithm gives an accuracy of 70\% for \textit{Walking} and \textit{Stationary} activities whereas it is less for the other two activities. We observed that the algorithm gets confused between these two activities. We suspect that the confusion is due to the fact that \textit{Splitting} and \textit{Approaching} are more abrupt in the motion dynamics than \textit{Walking} and \textit{Stationary}, which results in a poorer estimate of eigenvalues over the window of $L$ frames. In \textit{CUHK} dataset, since groups in most of the videos are walking, we obtain an accuracy of $85\%$. Some of the qualitative results on the videos from \textit{BEHAVE} and \textit{CUHK} dataset are given in Fig. \ref{fig:act_beh} and Fig. \ref{fig:act_cukh}, respectively.

\begin{figure*}
\centering
	\subfloat[\label{fig:confusion}]{\includegraphics[scale=0.355]{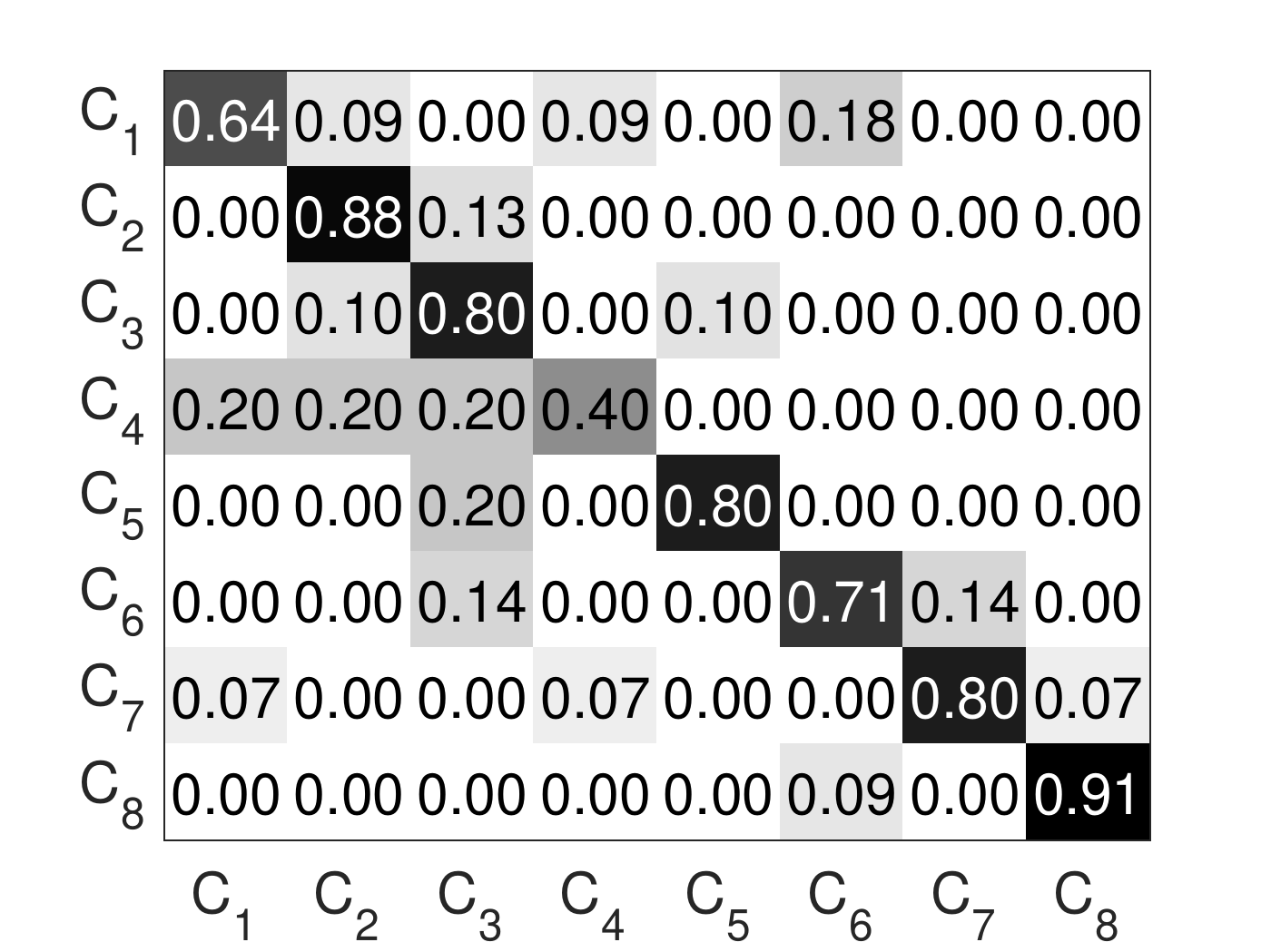}}
	\subfloat[\label{fig:OOB}]{\includegraphics[scale=0.355]{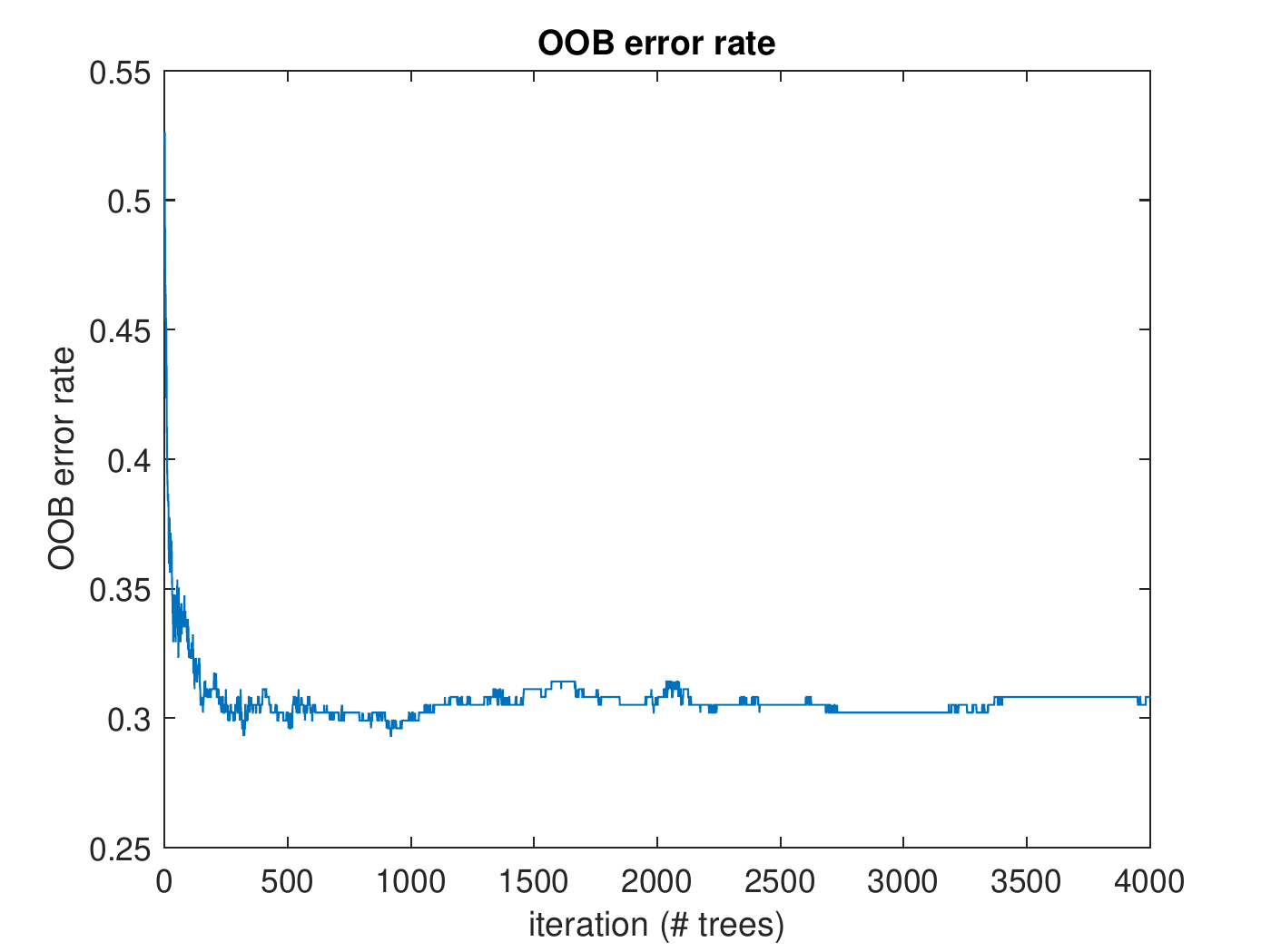}}
	\subfloat[\label{fig:Importance}]{\includegraphics[scale=0.355]{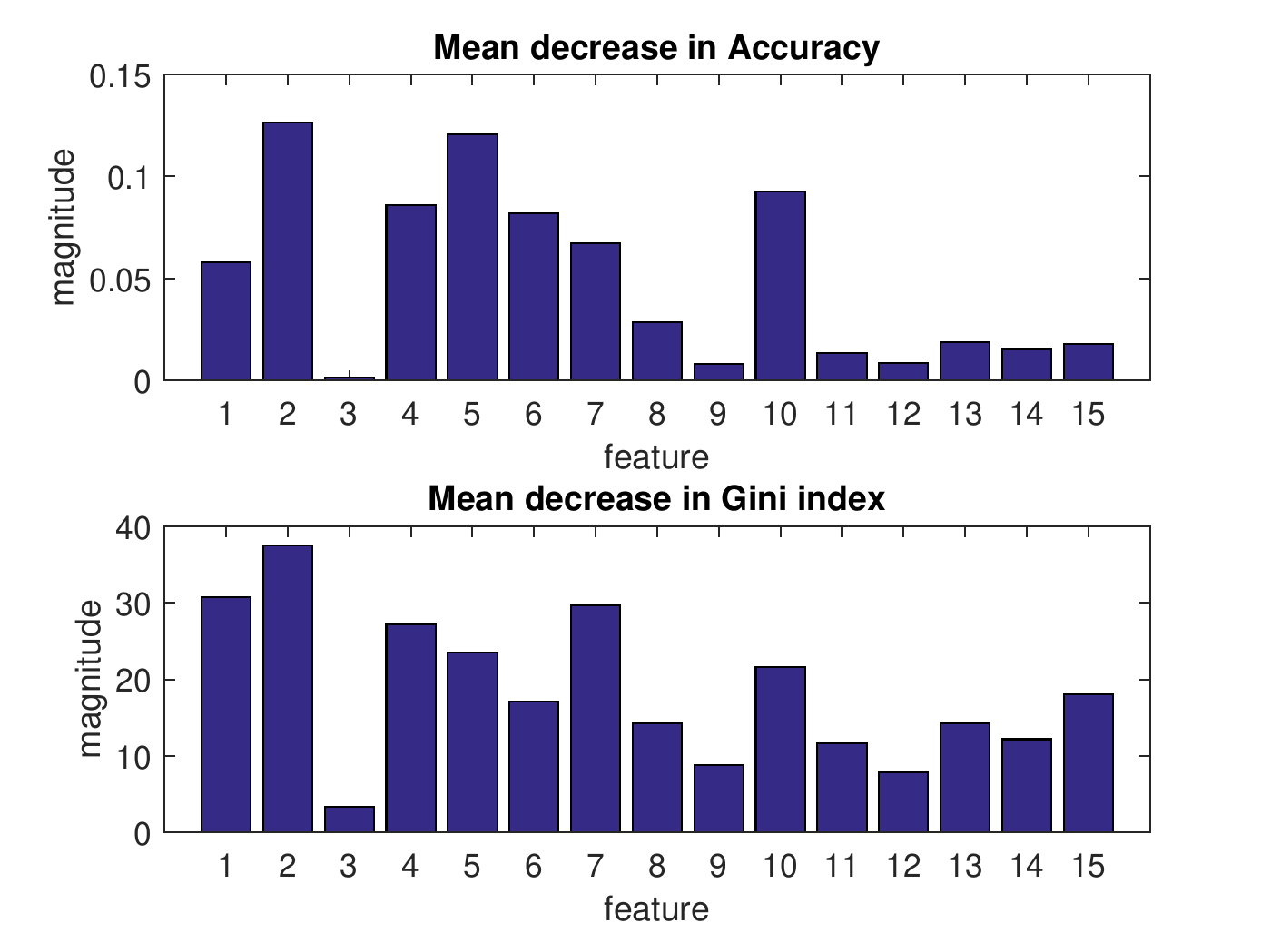}}
	\caption{(a) Confusion matrix with categories represented as C1 to C8 (true classes along the column and predicted classes along the row), (b) Out of bag (OOB) error, (c) Importance plot for the features}
\end{figure*}


\subsection{Crowd Video Classification}
Since we update the interaction model with each incoming frame as explained in Section $6$, we collect group level features at regular intervals. From each class, we randomly pick 70\% feature vectors to train the classifier and the remaining for testing. As discussed before, we use random forest as a classifier with $n=17$ and $m=4$. We run the classifier 100 times with random splits of dataset for training and testing. To avoid over-fitting, the training data and testing data do not contain features from the same video. The average accuracy obtained is around 74\%, an improvement over \cite{scene} where the reported accuracy is 70\%. The confusion matrix is shown in Fig.~\ref{fig:confusion}. From this figure, it is seen that classification of the crowd for Class 4 (Class Merge) is difficult, while the rest of the classes could be categorized quite easily using the proposed method. The OOB error, which indicates the generalized error, converges to a value 30\% as shown in Fig.~\ref{fig:OOB}. The importance plots, which show the significance of each group level feature in the classification are shown in Fig.~\ref{fig:Importance}. It shows that the group density and histogram of eigenvalues are important for classification. 

\section{Conclusions}
\label{conclusion}
In this work, we presented a framework for analysis of medium dense crowd videos at various levels. We proposed a first order dynamical system to model agent trajectories collectively and subsequently demonstrated the effectiveness of this interaction model for group detection. We also show how eigenvalues of the model characterize group activities. We then showed the effectiveness of group level features in crowd video classification. 

Our algorithm assumes the availability of tracks which itself is a challenge in many crowded videos due to occlusion and other tracking problems. As a next goal, we aspire to define a unified framework where the proposed model and a tracker work together to improve each other's  performance in crowded videos by incorporating group interaction cues.

\bibliographystyle{ieee} 
\bibliography{Paper}

\begin{thebibliography}{10}
\expandafter\ifx\csname url\endcsname\relax
  \def\url#1{\texttt{#1}}\fi
\expandafter\ifx\csname urlprefix\endcsname\relax\def\urlprefix{URL }\fi
\expandafter\ifx\csname href\endcsname\relax
  \def\href#1#2{#2} \def\path#1{#1}\fi

\bibitem{survey1}
T.~Li, H.~Chang, M.~Wang, B.~Ni, R.~Hong, S.~Yan, Crowded scene analysis: A
  survey, IEEE Transactions on Circuits and Systems for Video Technology 25~(3)
  (2015) 367--386.

\bibitem{survey2}
M.~Thida, Y.~L. Yong, P.~Climent-P{\'e}rez, H.-l. Eng, P.~Remagnino, A
  literature review on video analytics of crowded scenes, in: Intelligent
  Multimedia Surveillance, Springer, 2013, pp. 17--36.

\bibitem{survey3}
J.~S.~J. Junior, S.~Musse, C.~Jung, Crowd analysis using computer vision
  techniques, IEEE Signal Processing Magazine 5~(27) (2010) 66--77.

\bibitem{survey4}
B.~Zhan, D.~N. Monekosso, P.~Remagnino, S.~A. Velastin, L.-Q. Xu, Crowd
  analysis: a survey, Machine Vision and Applications 19~(5-6) (2008) 345--357.

\bibitem{behave}
S.~Blunsden, R.~Fisher, The behave video dataset: ground truthed video for
  multi-person behavior classification, Annals of the BMVA 4~(1-12) (2010) 4.

\bibitem{scene}
J.~Shao, C.~C. Loy, X.~Wang, Scene-independent group profiling in crowd, in:
  CVPR, 2014, IEEE, 2014, pp. 2227--2234.

\bibitem{vision}
W.~Ge, R.~T. Collins, R.~B. Ruback, Vision-based analysis of small groups in
  pedestrian crowds, IEEE Trans. PAMI 34~(5) (2012) 1003--1016.

\bibitem{spectral}
A.~Y. Ng, M.~I. Jordan, Y.~Weiss, et~al., On spectral clustering: Analysis and
  an algorithm, Advances in neural information processing systems 2 (2002)
  849--856.

\bibitem{mehran}
R.~Mehran, B.~E. Moore, M.~Shah, A streakline representation of flow in crowded
  scenes, in: Computer Vision--ECCV 2010, Springer, 2010, pp. 439--452.

\bibitem{solmaz}
B.~Solmaz, B.~E. Moore, M.~Shah, Identifying behaviors in crowd scenes using
  stability analysis for dynamical systems, IEEE Transactions on Pattern
  Analysis and Machine Intelligence 34~(10) (2012) 2064--2070.

\bibitem{motion}
Y.~Benabbas, N.~Ihaddadene, C.~Djeraba, Motion pattern extraction and event
  detection for automatic visual surveillance, Journal on Image and Video
  Processing 2011 (2011) 7.

\bibitem{dc}
W.~Lin, Y.~Mi, W.~Wang, J.~Wu, J.~Wang, T.~Mei, A diffusion and
  clustering-based approach for finding coherent motions and understanding
  crowd scenes, IEEE Transactions on Image Processing 25~(4) (2016) 1674--1687.

\bibitem{anamoly2}
A.~Pennisi, D.~D. Bloisi, L.~Iocchi, Online real-time crowd behavior detection
  in video sequences, Computer Vision and Image Understanding 144 (2016)
  166--176.

\bibitem{sfm1}
D.~Helbing, P.~Molnar, Social force model for pedestrian dynamics, Physical
  review E 51~(5) (1995) 4282.

\bibitem{sethi}
R.~J. Sethi, A.~K. Roy-Chowdhury, Individuals, groups, and crowds: Modelling
  complex, multi-object behaviour in phase space, in: IEEE International
  Conference on Computer Vision Workshops (ICCV Workshops), 2011, IEEE, 2011,
  pp. 1502--1509.

\bibitem{cf}
B.~Zhou, X.~Tang, X.~Wang, Coherent filtering: detecting coherent motions from
  crowd clutters, in: Computer Vision--ECCV 2012, Springer, 2012, pp. 857--871.

\bibitem{scs}
F.~Solera, S.~Calderara, R.~Cucchiara, Socially constrained structural learning
  for groups detection in crowd, IEEE transactions on pattern analysis and
  machine intelligence 38~(5) (2016) 995--1008.

\bibitem{Ge}
W.~Ge, R.~T. Collins, B.~Ruback, Automatically detecting the small group
  structure of a crowd, in: Workshop on Applications of Computer Vision (WACV),
  2009, IEEE, 2009, pp. 1--8.

\bibitem{soch}
J.~{\v{S}}ochman, D.~C. Hogg, Who knows who-inverting the social force model
  for finding groups, in: IEEE International Conference on Computer Vision
  Workshops (ICCV Workshops), 2011, IEEE, 2011, pp. 830--837.

\bibitem{sri}
V.~Srikrishnan, S.~Chaudhuri, Crowd motion analysis using linear cyclic
  pursuit, in: International Conference on Pattern Recognition (ICPR), 2010,
  IEEE, 2010, pp. 3340--3343.

\bibitem{L1G}
M.~Schmidt, G.~Fung, R.~Rosales, Optimization methods for l1-regularization,
  University of British Columbia, Technical Report TR-2009 19.

\bibitem{RF}
L.~Breiman, Random forests, Machine learning 45~(1) (2001) 5--32.

\bibitem{biwi}
S.~Pellegrini, A.~Ess, K.~Schindler, L.~Van~Gool, You'll never walk alone:
  Modeling social behavior for multi-target tracking, in: 2009 IEEE 12th
  International Conference on Computer Vision, IEEE, 2009, pp. 261--268.

\bibitem{cbe}
A.~Lerner, Y.~Chrysanthou, D.~Lischinski, Crowds by example, in: Computer
  Graphics Forum, Vol.~26, Wiley Online Library, 2007, pp. 655--664.

\bibitem{veiig}
S.~Bandini, A.~Gorrini, G.~Vizzari, Towards an integrated approach to crowd
  analysis and crowd synthesis: A case study and first results, Pattern
  Recognition Letters 44 (2014) 16--29.

\bibitem{nmi}
M.~Wu, B.~Sch{\"o}lkopf, A local learning approach for clustering, in: Advances
  in neural information processing systems, 2006, pp. 1529--1536.

\bibitem{purity}
C.~C. Aggarwal, A human-computer interactive method for projected clustering,
  IEEE Transactions on Knowledge and Data Engineering 16~(4) (2004) 448--460.

\bibitem{ri}
W.~M. Rand, Objective criteria for the evaluation of clustering methods,
  Journal of the American Statistical association 66~(336) (1971) 846--850.

\end{thebibliography}
\end{document}